\RequirePackage[T1]{fontenc}
\documentclass[letterpaper,journal]{IEEEtran}
\ifdefined\pdfminorversion
\fi
\ifdefined\XeTeXversion

\fi
\usepackage{amssymb}
\usepackage{amsmath,amsfonts}
\usepackage{algorithmic}
\usepackage{algorithm}
\usepackage{siunitx}
\usepackage{array}
\usepackage{textcomp}
\usepackage{stfloats}
\usepackage{url}
\usepackage{verbatim}
\usepackage{graphicx}
\usepackage{cite}
\usepackage{tabularx}
\usepackage{booktabs}
\usepackage[table,xcdraw]{xcolor}
\usepackage{multirow}
\usepackage{float}
\usepackage{makecell}
\usepackage{capt-of}

\newcommand{\tableguideline}{%
    \noalign{\vskip 1pt
        \hbox to \columnwidth{\leaders\hbox{\rule{1.5pt}{0.3pt}\hskip 1.5pt}\hfill}%
        \vskip 1pt}}

\begin{document}

\title{Physically Guided TeX Generation for \\ Low-Light Visual Tasks}

\author{Daoheng Li, Jianjun Qian, Mengkai Yan, and Jian Yang
\thanks{
Daoheng Li and Jianjun Qian are with the PCA Lab, School of Computer Science and Engineering, Nanjing University of Science and Technology, Nanjing, China (e-mail: lidaoheng@njust.edu.cn; csjqian@njust.edu.cn).
}%
\thanks{
Jian Yang is with the PCA Lab, School of Computer Science and Engineering, Nanjing University of Science and Technology, Nanjing, China, and also with the PCA Lab, School of Intelligence Science and Technology, Nanjing University, Nanjing, China (e-mail: csjyang@njust.edu.cn).
}%
\thanks{
Mengkai Yan is with the College of Computer and Software, Hohai University, Nanjing, China (e-mail: ymk@hhu.edu.cn).

}

}


\IEEEaftertitletext{\begin{minipage}{\textwidth}
    \centering
    \includegraphics[pagebox=cropbox,width=0.925\textwidth]{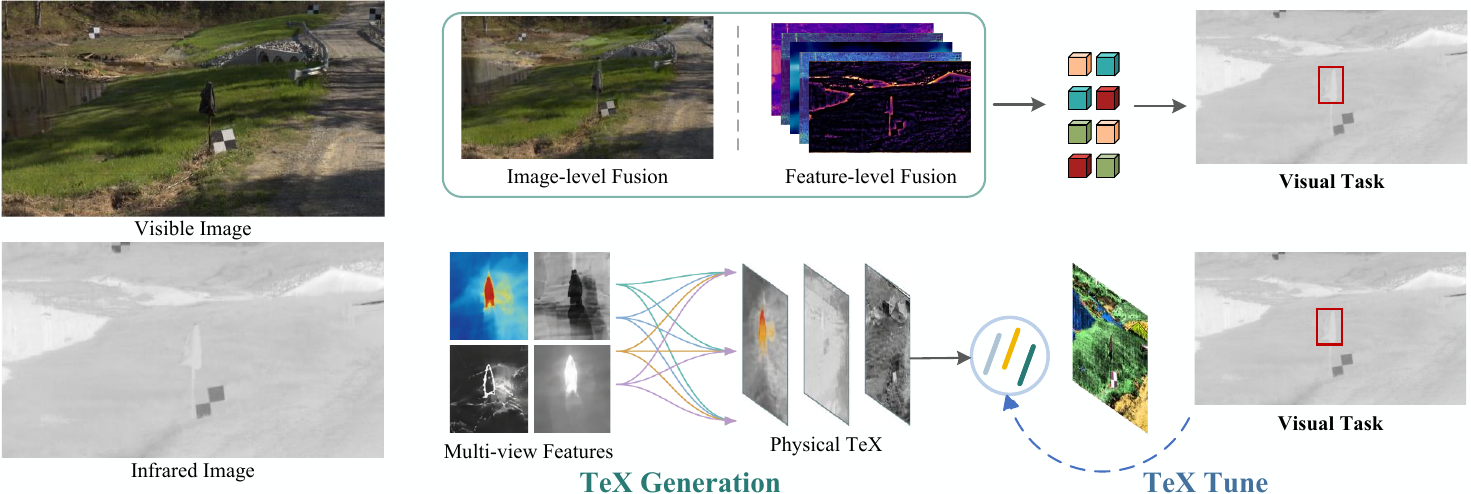}\\[-1pt]
    {\small (a) From fusion to TeX Generation and task-specific tuning.}\\[5pt]
    \includegraphics[pagebox=cropbox,width=0.925\textwidth]{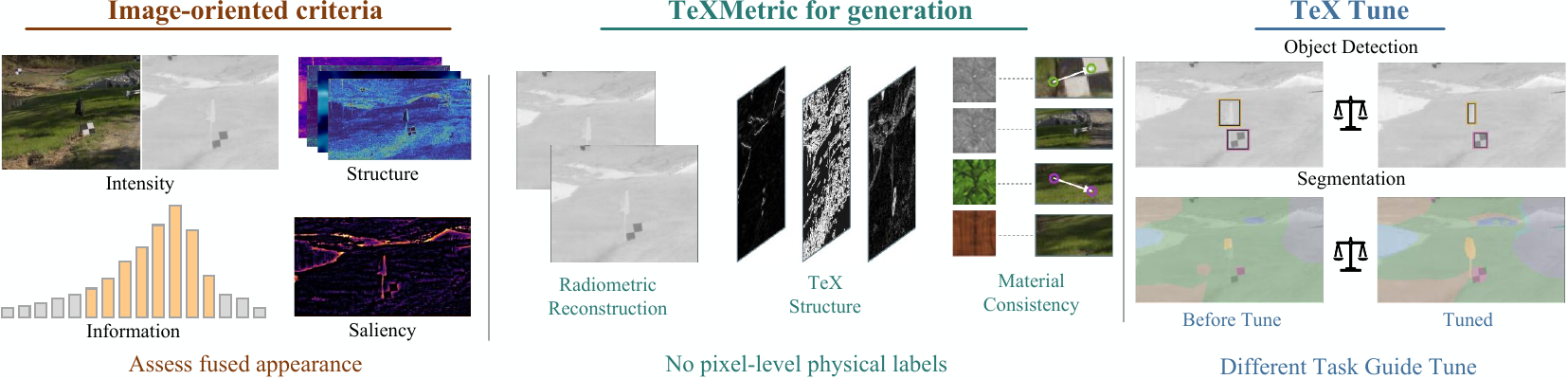}\\[-1pt]
    {\small (b) Image-oriented criteria, TeXMetric generation constraints, and TeX Tune.}
    \captionof{figure}{Motivation for TeX-Gen. (a) Appearance-based fusion versus physically guided TeX generation with task-specific tuning. (b) Image-oriented criteria, TeXMetric constraints, and task-guided channel tuning.}
    \label{Fig_Introduction_TeX_Overview}
\end{minipage}
}
\maketitle
\begin{abstract}
Low-light conditions reduce visible contrast and obscure details, while infrared images lack color and texture, motivating the fusion of infrared and visible information for visual tasks.
Existing fusion methods primarily optimize image appearance or task-specific features without explicitly disentangling the coupled factors underlying thermal observations, limiting their physical interpretation under thermal degeneracy.
Hyperspectral thermal measurements enable physical decomposition, but they require specialized imaging hardware, while pixel-level physical annotations for supervision or evaluation incur additional costs.
We propose TeX-Gen, a physically guided framework for TeX generation from paired visible and infrared images in low-light visual tasks. 
TeX Generation constructs complementary multi-view features and employs independent branch-specific gates to route cues into the Temperature, Emissivity, and Texture maps.
TeXMetric integrates radiometric reconstruction, channel structure, and material consistency to assess the generated maps without requiring pixel-level physical labels. 
TeX Tune improves performance through task-guided adjustments, jointly optimizing the adjustment modules and task network while keeping the generator frozen.
Experiments show that TeX-Gen achieves competitive task performance in low-light environments on the FLIR, LLVIP, M3FD, and MSRS datasets. 
Experiments on HADAR also examine the association between TeXMetric scores and downstream detection performance.
\end{abstract}


\section{Introduction}

Low-light conditions reduce visible contrast and obscure object boundaries, motivating the use of infrared sensing to retain scene information through emitted and reflected thermal radiation~\cite{chen2021infrared,jia2021llvip}.
Infrared images nevertheless lack color and often provide limited fine texture, which visible observations can complement when these cues remain available~\cite{bayoudh2022survey,zhang2023visible}.
The two modalities are therefore used together to improve performance on visual tasks under challenging illumination~\cite{zhang2021guided,qingyun2022cross,liu2022target,tang2022image}.
The central goal is to retain task-relevant information when visible cues weaken, rather than merely producing a brighter image.

Visible and infrared observations are commonly combined through feature-level or image-level fusion, as illustrated in Fig.~\ref{Fig_Introduction_TeX_Overview}(a).
Feature-level methods extract informative cues from both modalities and integrate them within a task network through learned interaction or attention~\cite{zhang2021guided,chen2022multimodal,qingyun2022cross,zeng2024mmi}.
Image-level methods instead produce a fused image that emphasizes observed targets while retaining complementary scene detail for a subsequent visual network.
Their objectives preserve source intensity and gradients~\cite{li2018densefuse,xu2020u2fusion}, reinforce structure~\cite{ma2022swinfusion,tang2022piafusion}, or improve visual realism~\cite{ma2019fusiongan,yue2023dif}.
However, continued improvement in image quality does not ensure continued gains in task accuracy.
TDHF reports that task accuracy can saturate or decline as visual-quality scores increase~\cite{li2025tdhf}.
Task-guided fusion introduces prediction signals to address this mismatch~\cite{tang2022image,zhao2023metafusion,dang2025ddfd,li2025tdhf}.
However, these representations remain coupled to a particular task objective, which can limit their reuse when downstream requirements change.

Beyond this limitation, appearance-based fusion does not explicitly distinguish the factors that produce thermal observations.
Thermal degeneracy further complicates this interpretation: different combinations of temperature and emissivity can produce similar thermal responses~\cite{bao2023heat,li2026mvrmoe}.
Physical decomposition therefore offers a more structured way to organize scene information than appearance alone.
Hyperspectral thermal images record radiative energy over narrow spectral bands, enabling TeXNet to estimate temperature $T$, emissivity $e$, and texture $X$ on HADAR~\cite{bao2023heat}.
Specialized sensors, demanding acquisition, and the preparation of pixel-level physical maps make this approach costly to deploy.
MVR-MoE extends TeX generation to visible and infrared pairs by combining multi-view structural representations with a mixture-of-experts design and radiometric reconstruction~\cite{li2026mvrmoe}, 
but its multi-view construction requires offline preprocessing and it lacks a dedicated criterion for evaluating the $T$, $e$, and $X$ maps.

To address these problems, we propose TeX-Gen, a framework that generates physically guided TeX maps for low-light visual tasks, as illustrated in Fig.~\ref{Fig_Introduction_TeX_Overview}(a).
TeX-Gen comprises TeX Generation, where a multimodal generator generates TeX under TeXMetric constraints, and TeX Tune, which provides task-specific tuning by adjusting the generated maps while keeping the generation mapping fixed.
The generator constructs four complementary feature views online, including cross-modal discrepancy, co-activation, spatial exchange, and channel attention. 
The $T$, $e$, and $X$ branches share the same architecture but maintain independent parameters and sample-conditioned HyperConnection gates, allowing each branch to route the shared feature views into channel-specific representations.
Infrared observations provide the radiometric reference, while visible cues complement structural and material analysis where they remain informative under low illumination.
Here, $T$, $e$, and $X$ denote the Temperature, Emissivity, and Texture maps, respectively.
TeXMetric combines radiometric reconstruction, channel structure, and material consistency to constrain generation without pixel-level physical labels, as illustrated in Fig.~\ref{Fig_Introduction_TeX_Overview}(b). 
TeX Tune makes slight task-specific adjustments to the generated TeX using multi-scale, local-activity, and region-aware cues. 
The adjustment modules and task network are jointly optimized while the generator remains frozen.

This paper is an extended version of our ACM Multimedia 2026 conference paper, MVR-MoE~\cite{li2026mvrmoe}.
The main extensions over the conference paper are as follows:
(1) TeX generation is redesigned by replacing offline multi-view preprocessing with online cross-modal feature construction and channel-specific HyperConnection routing;
(2) a TeX-specific evaluation metric, termed TeXMetric, is introduced to assess TeX generation;
(3) TeX Tune is introduced to make task-specific adjustments to the generated TeX while keeping the generator frozen.
To summarize, our key contributions are as follows:
\begin{itemize}
\item A physics-guided TeX generation framework is developed for low-light visual tasks. By combining infrared radiometric information with complementary visible cues, it generates channel-specific Temperature, Emissivity, and Texture maps for downstream tasks.
\item We introduce TeXMetric to constrain and evaluate TeX generation. Radiometric reconstruction, channel structure, and material consistency are jointly considered, enabling TeX generation to be constrained and evaluated without pixel-level physical labels.
\item TeX Tune provides a way to tune common TeX maps to different downstream tasks through slight task-specific adjustments. Its effectiveness is evaluated on object detection and semantic segmentation across multiple datasets.
\end{itemize}

\section{Related Work}

\subsection{Visible and Infrared Fusion}
Visible and infrared fusion integrates complementary sensor observations at the feature or image level.
The two routes differ in their feature representation rather than their input sensing modalities.
Feature-level fusion embeds the two sensing modalities in a common task network, allowing visual-task supervision to shape cross-modal features directly.
Depending on the interaction depth, feature-level methods combine shallow appearance cues, exchange intermediate features, or aggregate modality-specific predictions near the task head. 
Mid-level fusion retains separate encoders while integrating semantically aligned features, whereas recurrent and cyclic designs progressively exchange information between the two streams~\cite{dhanaraj2020vehicle,zhang2020multispectral}.
Spatial and channel attention further estimate where and how strongly each modality should contribute, which is useful when visible texture degrades under poor illumination or infrared contrast weakens in thermally ambiguous regions~\cite{zhang2021guided,cao2023multimodal}.
Decision-level methods instead retain independent predictions and combine their confidence distributions, limiting premature interference between heterogeneous features~\cite{chen2022multimodal}.
Paired observations can differ around object boundaries because of occlusion, saturation, or registration errors. Long-range attention and conflict-aware learning distinguish complementary evidence from these cross-modal discrepancies~\cite{fu2023lraf,he2023multispectral}.
Iterative cross-attention repeatedly exchanges modality information, while frequency decomposition separates structural content from high-frequency details before fusion~\cite{shen2024icafusion,li2025fd2}.
Transformer and Mamba variants extend cross-modal interaction through multidimensional attention, infrared-centered aggregation, state-space propagation, and wavelet-domain exchange~\cite{yang2024multidimensional,zeng2024mmi,dong2024fusion,hwang2025multispectral,zhu2025wavemamba}.
Recent methods also address specific sources of multimodal features. 
RegisterBridgeMM focuses on registration, InfraNet regulates visible guidance according to modality quality, distribution-aligned fusion reduces cross-modal distribution mismatch, and HMAL learns modality advantages across feature hierarchies~\cite{wang2026registerbridgemm,feng2026infranet,hao2026distribution,xiang2026hmal}.
Frequency- and wavelet-based methods exchange complementary evidence, while infrared-centric and cross-hierarchical transformers employ asymmetric attention and gated feature interaction~\cite{wu2026fredft,zhu2025wavemamba,hwang2025multispectral,hou2026chudetr}.
Illumination-aware and uncertainty-aware methods further account for lighting variation and modality uncertainty~\cite{lu2025illumination,sun2022drone}.

Image-level fusion integrates complementary visible and infrared observations into a single representation compatible with conventional vision networks.
Decomposition-based approaches separate source images into multiple scales, sparse components, gradients, or low-rank and detail layers and aggregate the components using predefined activity measures~\cite{liu2015general,ma2016infrared,li2020mdlatlrr}.
Deep models replace hand-designed rules with learned encoders, decoders, and fusion operators.
DenseFuse aggregates densely connected features, FusionGAN uses adversarial learning to balance thermal saliency and visible texture, and U2Fusion learns a unified unsupervised objective across fusion settings~\cite{li2018densefuse,ma2019fusiongan,xu2020u2fusion}.
Illumination-aware methods vary the fusion preference across lighting conditions, thereby suppressing unreliable visible cues in dark regions while retaining informative texture elsewhere~\cite{tang2022piafusion}.
Transformer-based methods employ self-attention or cross-attention to model nonlocal relationships and exchange information across spatially distant regions~\cite{ma2022swinfusion,tang2023datfuse,tang2024itfuse}.
Frequency-domain networks explicitly separate low-frequency intensity from high-frequency detail, while diffusion and rectified-flow models describe fusion as a conditional generative process~\cite{wang2024efficient,yue2023dif,wang2025rffusion}.
Training criteria in this category commonly measure intensity retention~\cite{li2018densefuse,xu2020u2fusion}, gradients and structural similarity~\cite{ma2022swinfusion,tang2022piafusion}, or adversarial realism~\cite{ma2019fusiongan}.
Such criteria are appropriate for assessing a single visually coherent image, but they do not determine whether separate channels follow the temperature, emissivity, and environmental-radiation factors of thermal image formation.
Both routes organize complementary visual evidence without explicitly assigning thermal image-formation factors to separate channels.
Feature fusion is often tied to a task architecture, while image-quality objectives alone do not determine performance.

\subsection{Task-Guided Fusion}
Task-guided fusion addresses the mismatch that arises when a fusion model is optimized with image-quality criteria and then evaluated by a visual-task network with a different objective.
One strategy trains fusion and high-level prediction jointly, so downstream gradients directly influence the fusion process.
Another strategy feeds task signals back to the fusion network through auxiliary branches, meta-features, or an alternating optimization loop.
Target-aware dual adversarial learning strengthens object-related regions during image synthesis, while semantic-aware fusion places high-level vision inside the fusion loop~\cite{liu2022target,tang2022image}.
MetaFusion embeds task-derived meta-features into the fusion process, connecting the generated image with the behavior of the task model~\cite{zhao2023metafusion}.
Multi-interactive learning jointly exchanges fusion and high-level semantic cues so that region semantics and image details can support each other~\cite{liu2023multi}.
Diffusion-based denoising and closed-loop dynamic networks further adjust fusion behavior according to task requirements or task-dependent feedback~\cite{dang2025ddfd,yang_customizedfusion}.
TDHF separates hierarchical fusion from task-driven adjustment and reports that stronger image-quality scores need not yield higher downstream accuracy~\cite{li2025tdhf}.
These studies demonstrate the benefits of task feedback, while their primary results remain fused images or shared feature tensors rather than channels with explicitly assigned physical roles.
This task-dependent optimization limits the reuse of learned fusion representations across downstream objectives.
These limitations motivate generating a common physical-channel representation and tuning its responses for different visual tasks without changing the generator.

\subsection{Physically Motivated Infrared Analysis}
Physically motivated methods organize infrared information according to image formation rather than visual saliency alone.
Thermal measurements integrate wavelength-dependent radiance according to the sensor response and depend on object temperature, spectral emissivity, reflected environmental radiation, and atmospheric effects~\cite{bao2023heat}.
Hyperspectral thermal imaging records energy in narrow spectral bands to help separate these coupled factors.
TeXNet estimates temperature, spectral emissivity, and texture from the hyperspectral measurements in HADAR~\cite{bao2023heat}.
Its texture channel reflects geometrical and environmental thermal structure.
$X$ is parameterized by the normalized mixing coefficient $V$. 
In the HADAR setting, $V$ is the normalized code visualized for comparison with the available channel labels, whereas TeXNet estimates texture from hyperspectral measurements.
Related physical decompositions address other acquisition settings.
VT-Intrinsic uses a visible-thermal pair to separate reflectance and shading, whereas PhysIR-Splat incorporates thermal radiative transfer into three-dimensional Gaussian splatting~\cite{yuan_vtintrinsic,gao_physirsplat}.
PhyFusion introduces modality-specific physical priors into visible-infrared image fusion rather than relying only on generic appearance constraints~\cite{jiang_phyfusion}.
MVR-MoE combines multi-view structural representations with a mixture-of-experts design to generate TeX from paired visible and infrared images~\cite{li2026mvrmoe}.
Broadband radiometric reconstruction links the generated channels to thermal image formation without hyperspectral acquisition. 
However, reconstruction alone does not fully constrain the structural and material relationships among the channels, motivating additional channel-specific constraints.

\section{Method}

\subsection{Overview}
\begin{figure*}[t]
    \centering
    \includegraphics[pagebox=cropbox,width=0.925\textwidth]{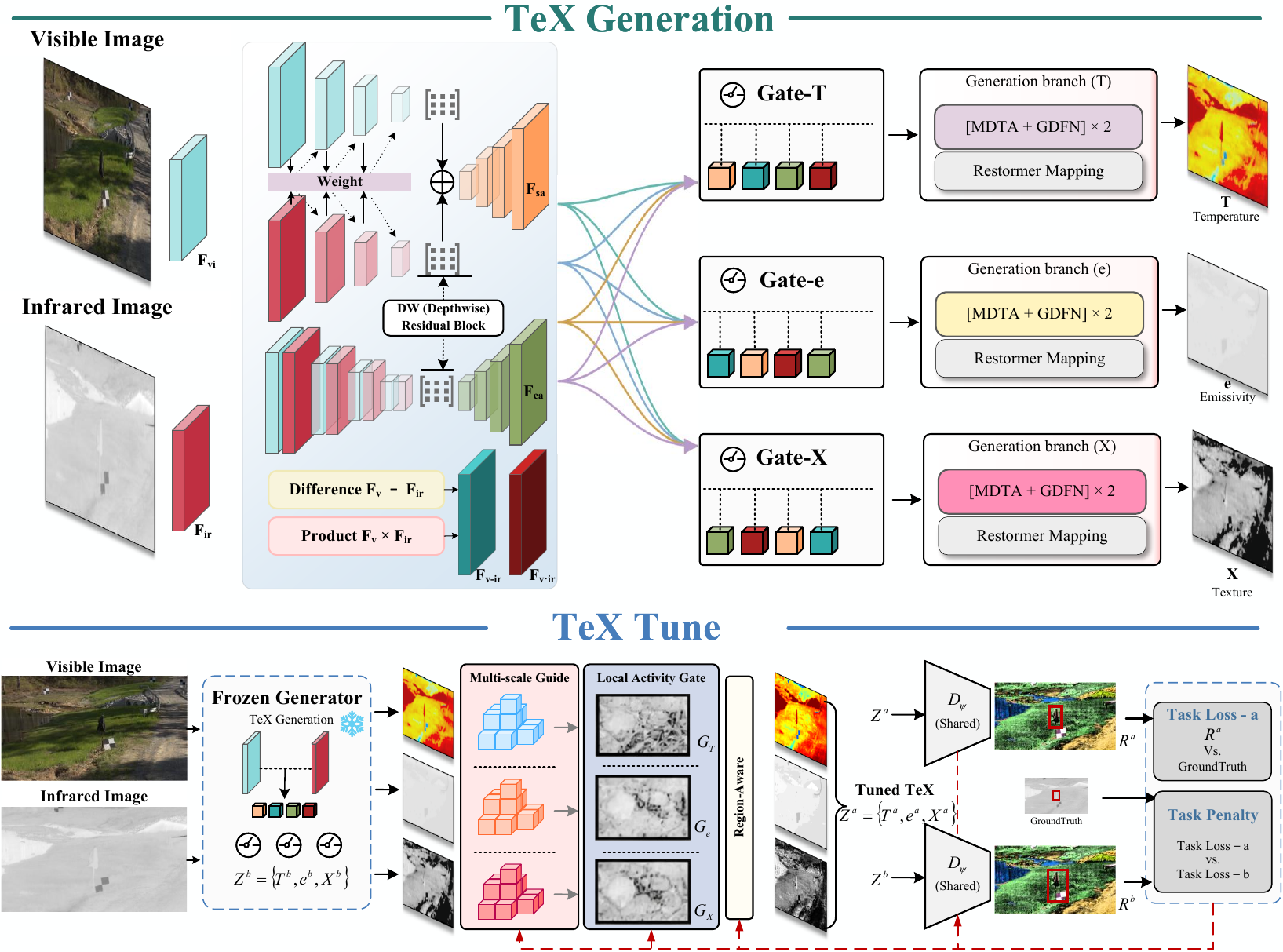}
    \caption{
    Overview of TeX-Gen. TeX Generation constructs multi-view features online and selects complementary evidence through channel-specific gates to form the baseline TeX $Z^b$ under TeXMetric constraints. 
    With the generator frozen, TeX Tune applies task-specific corrections to $Z^b$ through Multi-scale Guide, Local Activity Gate, and Region-Aware processing before passing the tuned TeX $Z^a$ to the complete downstream task network. 
    Task supervision jointly optimizes the channel-tuning modules and task network, whereas the task penalty updates only the channel-tuning modules.}
    \label{Fig_Method_Framework_Overview}
\end{figure*}

TeX-Gen generates physically guided TeX maps from infrared and visible images in low-light conditions, with task-specific tuning to adapt TeX to different downstream tasks, as illustrated in Fig.~\ref{Fig_Method_Framework_Overview}.
The framework comprises TeX generation and task-specific tuning of the generated TeX maps.
The former combines the TeX Generation network with TeXMetric, which constrains and assesses TeX. 
The latter is implemented by TeX Tune, which adjusts the generated TeX while keeping the generator frozen.
The paired images are encoded into aligned features $F_v$ and $F_{ir}$.
Their difference $F_v-F_{ir}$ exposes disagreement and their element-wise product $F_v\odot F_{ir}$ captures co-activation.
Alternating cross-modal exchange forms the spatial-attention feature $F_{sa}$, while channel concatenation followed by channel-aware processing forms the channel-attention feature $F_{ca}$.
All four views are generated online within the generator.
Each generation branch uses a dedicated sample-conditioned HyperConnection gate to select features from these views before decoding $TeX^{b}=\operatorname{Concat}(Z_T^b,Z_e^b,Z_X^b)\in[0,1]^{3\times H\times W}$.
TeXMetric combines radiometric reconstruction, channel structure, and material consistency to constrain generation without pixel-level physical labels. 
During TeX Tune, the generator remains frozen, while three independent adjustment branches apply task-specific residual corrections to the generated channels. 
The adjustment modules and downstream task network are jointly optimized using task supervision. An additional relative penalty updates only the adjustment modules when the tuned task loss exceeds the detached baseline loss.
TeX Generation does not use downstream task losses.

\subsection{TeX Generation}
\subsubsection{Multi-view Feature Construction}
For the aligned visible image $I_v$ and infrared image $I_{ir}$, the encoders $E_v$ and $E_{ir}$ produce equally sized latent maps,
\begin{equation}
    F_v=E_v(I_v),\; F_{ir}=E_{ir}(I_{ir}),\;
    F_v,F_{ir}\in\mathbb{R}^{C\times h\times w}.
\end{equation}
Here, $C$, $h$, and $w$ denote the feature channel count, height, and width, respectively, and $\mathbb{R}$ denotes the real numbers.
Four complementary feature views are extracted from the encoded pair,
$\mathcal{S}=\{F_{sa},F_{ca},F_{\Delta},F_{\mathrm{co}}\}$.

The discrepancy and co-activation views are denoted by $F_{\Delta}$ and $F_{\mathrm{co}}$, respectively:
\begin{equation}
    F_{\Delta}=F_v-F_{ir},\qquad
    F_{\mathrm{co}}=F_v\odot F_{ir},
\end{equation}
Here, $-$ and $\odot$ denote element-wise subtraction and multiplication, respectively.
The resulting features capture cross-modal discrepancy and co-activation, providing complementary inputs for the channel-specific HyperConnection gates.

The spatial-attention modality uses an alternating exchange of complementary information between the visible and infrared streams.
Let $H_v^{(0)}=F_v$ and $H_{ir}^{(0)}=F_{ir}$, and let $\mathcal{A}_v^{(j)}$ and $\mathcal{A}_{ir}^{(j)}$ denote spatial-attention transformations at step $j$.
The initial step processes each stream independently,
\begin{equation}
H_v^{(1)}=\mathcal{A}_v^{(1)}(H_v^{(0)}),\qquad
H_{ir}^{(1)}=\mathcal{A}_{ir}^{(1)}(H_{ir}^{(0)}).
\end{equation}
Each stream then receives a weighted contribution from the preceding state of the other stream:
\begin{equation}
\begin{aligned}
H_v^{(2)}&=\mathcal{A}_v^{(2)}\!\left(H_v^{(1)}+
W_{ir\rightarrow v}^{(1)}\odot H_{ir}^{(1)}\right),\\
H_{ir}^{(2)}&=\mathcal{A}_{ir}^{(2)}\!\left(H_{ir}^{(1)}+
W_{v\rightarrow ir}^{(1)}\odot H_v^{(1)}\right).
\end{aligned}
\end{equation}
The learned weights $W_{ir\rightarrow v}^{(1)}$ and $W_{v\rightarrow ir}^{(1)}$ control transfer from infrared to visible and from visible to infrared, respectively, with dimensions aligned to the features they multiply.
Parenthesized superscripts index exchange steps.
The final features are added and decoded by transposed convolution,
\begin{equation}
F_{sa}=\operatorname{Deconv}\left(H_v^{(J)}+H_{ir}^{(J)}\right),
\end{equation}
where $J$ denotes the terminal exchange step and $\operatorname{Deconv}$ denotes transposed convolution.
This alternating process allows visible edges and infrared thermal saliency to influence both streams before their spatial fusion.

The channel-attention modality concatenates the two encodings along the channel axis,
\begin{equation}
\begin{aligned}
    U_{ca}&=\operatorname{Conv}_{ca}\left(\operatorname{Concat}(F_v,F_{ir})\right),\\
    F_{ca}&=\mathcal{A}_{ca}(U_{ca}),
\end{aligned}
\end{equation}
Here, $U_{ca}$ is the intermediate feature produced by the learned convolution $\operatorname{Conv}_{ca}$, and $\mathcal{A}_{ca}$ performs channel-aware convolutional processing.
Concatenation preserves both sensor streams before learned channel mixing and reweighting.
All four constructed modalities have dimensions $C\times h\times w$.

\subsubsection{Channel-Specific HyperConnection Gates}
The three generation branches, which share the same architecture, have independent parameters and dedicated HyperConnection gates~\cite{zhu2024hyper}, indexed by $k\in\{T,e,X\}$, to govern the contribution of the constructed multi-view features:
\begin{equation}
    F_{\mathrm{use},k}=\mathcal{H}_{\mathrm{HC},k}
       \left(F_{sa},F_{ca},F_{\Delta},F_{\mathrm{co}}\right).
\end{equation}
Here $\mathcal{H}_{\mathrm{HC},k}$ denotes the dedicated gate for channel k.
The three gates have independent parameters and use sample-dependent connection weights to route the four shared feature views into channel-specific representations.

\subsubsection{Parallel TeX Generation}
The routed feature $F_{\mathrm{use},k}$ is delivered to its corresponding generation branch indexed by $k\in\{T,e,X\}$.
These branches have the same architecture and independent parameters.
Each branch stacks two Restormer-style transformer blocks~\cite{zamir2022restormer}, each combining Multi-Dconv Head Transposed Attention (MDTA) with a Gated-Dconv Feed-Forward Network (GDFN), followed by a channel-specific mapping head.
With $U_k^{(0)}=F_{\mathrm{use},k}$, the $j$-th block computes
\begin{equation}
\begin{aligned}
    \bar{U}_k^{(j)}&=U_k^{(j-1)}+
       \operatorname{MDTA}_{k,j}\left(\operatorname{LN}(U_k^{(j-1)})\right),\\
    U_k^{(j)}&=\bar{U}_k^{(j)}+
       \operatorname{GDFN}_{k,j}\left(\operatorname{LN}(\bar{U}_k^{(j)})\right),
       \quad j\in\{1,2\}.
\end{aligned}
\end{equation}
Here, $j$ indexes the two Restormer blocks, and $\operatorname{LN}$ denotes layer normalization.

Multi-Dconv Head Transposed Attention (MDTA) forms queries, keys, and values through pointwise and depthwise convolutions.
Within each attention head, its tensor is arranged as a channel-by-position matrix.
One attention head computes
\begin{equation}
    A=\operatorname{softmax}\left(\tau\bar{Q}\bar{K}^{\top}\right),
    \qquad O=AU_{\rm val},
\end{equation}
where $\bar{Q}$ and $\bar{K}$ are queries and keys normalized over spatial positions, $U_{\rm val}$ contains the value features, and $\tau$ is a learned scale.
The superscript $\top$ denotes matrix transpose. 
The row-wise softmax normalizes attention scores into weights. 
Here, $A$ is the resulting channel-attention matrix and $O$ is its feature matrix.
The resulting attention matrix models channel relationships from spatially aggregated evidence.
The heads are concatenated and projected back to the branch feature dimension.

The Gated-Dconv Feed-Forward Network (GDFN) expands the channel dimension with a pointwise convolution and applies a depthwise convolution.
Splitting the channels into $U_1$ and $U_2$ gives the gated transform
\begin{equation}
    \operatorname{GDFN}(U)=W_o
       \left(\operatorname{GELU}(U_1)\odot U_2\right).
\end{equation}
Here, $U$ is the block input, $U_1$ and $U_2$ are the two feature groups after expansion and local filtering, $\operatorname{GELU}$ is the Gaussian error linear unit activation, and $W_o$ is the learned projection.
This operation combines local filtering with content-dependent feature suppression.
The branch-specific Restormer mapping head then converts the final feature into one channel at the input image resolution in $[0,1]$,
\begin{equation}
    Z_k^b=\mathcal{M}_k(U_k^{(2)}),\qquad
    Z^b=\operatorname{Concat}(Z_T^b,Z_e^b,Z_X^b).
\end{equation}
Here, $\mathcal{M}_k$ is the mapping head of channel $k$, and $Z^b$ is the concatenated baseline TeX.
The independent parameters allow the three branches to specialize under the coupled TeXMetric objective.

\subsection{TeXMetric Constraints}
TeXMetric combines radiometric reconstruction, channel structure, and material consistency into a unified generation objective without pixel-level physical labels. 
Unlike generic image-quality criteria, it evaluates the physical relationships among the generated channels.

\subsubsection{Radiometric Reconstruction}
\begin{figure}[!t]
    \centering
    \includegraphics[pagebox=cropbox,width=0.77\columnwidth]{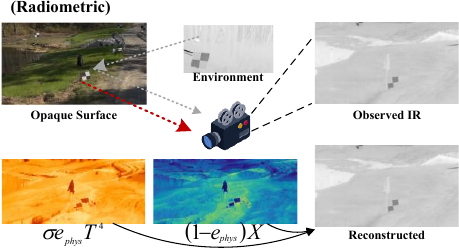}
    \caption{Radiometric reconstruction in TeXMetric. 
    Emitted and reflected environmental contributions form a normalized infrared reconstruction for comparison with the observed image.}
    \label{Fig_Method_TeXMetric_Radiometric}
\end{figure}

The reconstruction term is motivated by thermal image formation and operates on normalized infrared observations.
For a sensor response $R(\lambda)$ over $[\lambda_1,\lambda_2]$, a band measurement can be expressed as
\begin{equation}
    L_{ir}=\int_{\lambda_1}^{\lambda_2}R(\lambda)L_\lambda\,d\lambda.
\end{equation}
Here, $\lambda$ denotes wavelength, $\lambda_1$ and $\lambda_2$ bound the sensor band, and $L_{ir}$ is the band-integrated measurement.
The spectral radiance $L_\lambda=e_\lambda B(\lambda,T)+(1-e_\lambda)X_\lambda$ combines Planck radiance $B(\lambda,T)$ at temperature $T$, spectral emissivity $e_\lambda$, and environmental radiance $X_\lambda$ reflected by an opaque surface.
The implementation uses a normalized bolometric formulation to constrain the three generated channels, as illustrated in Fig.~\ref{Fig_Method_TeXMetric_Radiometric}.
The normalized Temperature is mapped to $T=T_{\min}+(T_{\max}-T_{\min})Z_T^b$, and the Emissivity map is decoded as $e_{\rm phys}=\mathcal{D}_e(Z_e^b)$.
The bounds $T_{\min}$ and $T_{\max}$ are in kelvin, and $\mathcal D_e$ maps the normalized Emissivity code to the emissivity used in reconstruction.
For $K$ fixed emissivity intervals, $l_m$ and $u_m$ denote the lower and upper bounds of interval $m$.
Let $a_m=(l_m+u_m)/2$ and $c_m=(m-\tfrac12)/K$, with $m=1,\ldots,K$.
The decoder is
\begin{equation}
\begin{aligned}
    p_m(z)=\frac{\exp(-|z-c_m|/\tau_e)}
       {\sum_{j=1}^{K}\exp(-|z-c_j|/\tau_e)},
    \mathcal{D}_e(z)=\sum_{m=1}^{K}p_m(z)a_m.
\end{aligned}
\end{equation}
Here, $z$ is a scalar normalized Emissivity code, $c_m$ is its code-space interval center, $a_m$ is the corresponding emissivity-interval midpoint, and $p_m(z)$ is its nonnegative assignment weight, with $\sum_m p_m(z)=1$.
The index $j$ in the denominator runs over the same $K$ intervals, and $\tau_e>0$ controls the smoothness of the assignment weights.
The ordered intervals provide an emissivity prior for the Emissivity map.
The HADAR dataset experiment instead outputs an emissivity estimate directly and uses $\mathcal{D}_e(z)=\operatorname{clip}(z,0,1)$, where $\operatorname{clip}(z,0,1)=\min(1,\max(0,z))$ acts elementwise.
Texture stores the normalized code $Z_X^b \to V$. 
This code parameterizes the texture representation $X$, which encodes spatial variation associated with environmental radiation. 
The infrared observation is reconstructed as
\begin{equation}
    Q_{ir}=\sigma e_{\rm phys}T^4+(1-e_{\rm phys})X.
\end{equation}
Following HADAR's formulation of environmental radiation through thermal lighting factors~\cite{bao2023heat}, $X$ is computed from $V$ as a weighted combination of the radiation contributions associated with two fixed effective environmental temperatures, $T_1$ and $T_2$. 
Using the Stefan--Boltzmann relation, we obtain
\begin{equation}
    X=\sigma\left(VT_1^4+(1-V)T_2^4\right),
\end{equation}
where $\sigma$ is the Stefan-Boltzmann constant, $T_1$ and $T_2$ are in kelvin, and $V\in[0,1]$ and $1-V$ are the respective mixing weights.
Here, $Q_{ir}$ models the reconstructed bolometric infrared observation by combining object-emitted radiation with the reflected environmental-radiation contribution. 
All products and fourth powers in these image-valued relations are evaluated pixelwise.
For the temperature bounds $T_{\min},T_{\max}$, the reconstructed normalized image is
\begin{equation}
    \tilde I_{ir}=
    \frac{Q_{ir}-Q_{\min}}{Q_{\max}-Q_{\min}},
\end{equation}
where $Q_{\min}=\sigma T_{\min}^4$ and $Q_{\max}=\sigma T_{\max}^4$.
The decoded channels provide consistency with the observed infrared image within this formulation.

The radiometric loss compares the reconstructed and measured infrared images,
\begin{equation}
\begin{aligned}
    \mathcal{L}_{\mathrm{rad}}={}\lambda_{\mathrm{rmse}}\sqrt{E_{\mathrm{rec}}+\varepsilon}
       +\lambda_{\mathrm{dist}}D_{\mathrm{KL}}(p(I_{ir})\|p(\tilde I_{ir})).
\end{aligned}
\end{equation}
Let $\Omega$ be the image pixel grid, $N=HW$ its size, and $B$ the batch size.
Here, $E_{\mathrm{rec}}=\frac{1}{BN}\sum_{n=1}^{B}\sum_{p\in\Omega}(\tilde I_{ir,n}(p)-I_{ir,n}(p))^2$ averages squared residuals over the batch and pixels.
For each image, $p(I)(p)=\exp(I(p))/\sum_{q\in\Omega}\exp(I(q))$ is the spatial intensity distribution.
The image index $n$ runs over the batch, $p$ and $q$ index pixel locations, and $p(I)$ denotes the spatial intensity distribution.
For two such distributions $P$ and $Q$, $D_{\mathrm{KL}}(P\|Q)=\sum_{p\in\Omega}P(p)\log(P(p)/Q(p))$ is their Kullback-Leibler divergence.
The weights $\lambda_{\rm rmse}$ and $\lambda_{\rm dist}$ balance the RMSE and spatial-distribution terms, respectively.
The RMSE term takes the square root after this batch-and-pixel average, whereas KL divergence is computed per image and averaged over the batch.
A small positive $\varepsilon$ ensures numerical stability.

\subsubsection{Channel Structure}
\begin{figure}[!t]
    \centering
    \includegraphics[pagebox=cropbox,width=0.75\columnwidth]{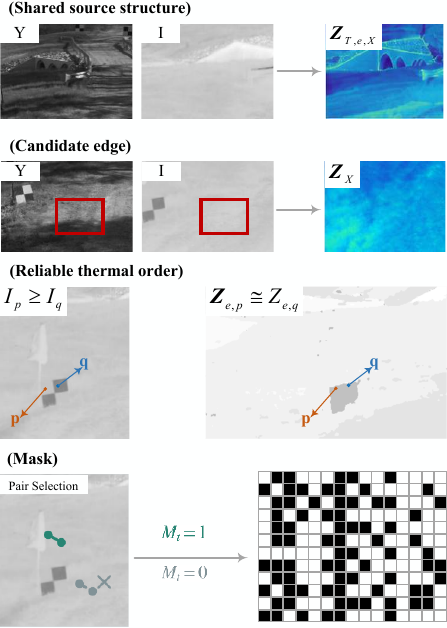}
    \caption{Channel-structure constraints involving all three channels. 
    Shared source structure guides $T$, $e$, and $X$. Candidate material edges are attenuated in $z_X$, while eligible pairs with similar $z_e$ guide temperature ordering in $z_T$.}
    \label{Fig_Method_TeXMetric_Structural}
\end{figure}

The three maps have complementary structural roles rather than identical boundaries (Fig.~\ref{Fig_Method_TeXMetric_Structural}).
For compactness, write $z_k=Z_k^b$ and $I=I_{ir}$ below, with visible luminance $Y=0.299R+0.587G+0.114B$.
Here, $R$, $G$, and $B$ are the visible red, green, and blue image channels.
All image values are clipped to $[0,1]$.
Binary masks select eligible pixels or pixel pairs and are detached from gradient computation.
Each mask equals one where its listed conditions hold and zero elsewhere, so masked losses accumulate over the selected locations.
Pixel pairs $(p,q)$ use the six offsets
$\mathcal{O}=\{(0,1),(1,0),(0,2),(2,0),(1,1),(1,-1)\}$.
For offset $o$, let $\mathcal{P}_o$ contain all pairs with $q=p+o$ and valid coordinates in the current batch.
Their relative difference is
\begin{equation}
    d_f(p,q)=\frac{f(p)-f(q)}{\max(|f(p)|,|f(q)|,\varepsilon_d)}.
\end{equation}
Here, $f$ is any scalar image map, and $\varepsilon_d>0$ stabilizes relative differences.
The signed quantity $d_f(p,q)$ is positive when $f(p)>f(q)$ and negative when the order is reversed. Its magnitude measures a relative rather than absolute intensity difference.
Subsequent pairwise formulas suppress $(p,q)$ in $d_f$ when the pair is clear.
For a pair mask $M$, its eligible-pair count is $N_M^o=\sum_{(p,q)\in\mathcal{P}_o}M(p,q)$.
Let $\mathcal{O}_M=\{o\in\mathcal{O}:N_M^o>0\}$ contain the offsets with eligible pairs.
Each pairwise loss first averages within an eligible offset and then averages equally over $\mathcal{O}_M$.
Thus, $|\mathcal O_M|$ counts nonempty offsets, not pairs, the outer average gives each such offset equal weight regardless of its eligible-pair count.

Shared ordinal structure is selected by
\begin{equation}
    M_s^{\rm ord}(p,q)=
    \begin{cases}
    1,&\scriptstyle |d_Y|>\tau_s,\ |d_I|>\tau_s,\ d_Yd_I>0,\\
    0,&\text{otherwise},
    \end{cases}
\end{equation}
where $\tau_s$ excludes source differences.
Both relative differences must exceed $\tau_s$ in magnitude, and the numerical product $d_Yd_I>0$ requires matching signs: the same endpoint must be brighter in both sources.
The corresponding shading-order term is
\begin{equation}
\begin{aligned}
    \mathcal{L}_{\rm shade}
    &=\frac{1}{\max(1,|\mathcal{O}_s|)}
      \sum_{o\in\mathcal{O}_s}\frac{1}{N_s^o}
      \sum_{(p,q)\in\mathcal{P}_o}M_s^{\rm ord}(p,q)\\
    &\qquad\cdot\max\left(0,\delta_s-d_{z_X}\operatorname{sign}(d_Y)\right),
\end{aligned}
\end{equation}
where $N_s^o=N_{M_s^{\rm ord}}^o$, $\mathcal{O}_s=\mathcal{O}_{M_s^{\rm ord}}$, and $\delta_s$ is the ordering margin.
For an eligible pair with $Y(p)>Y(q)$, zero penalty requires $d_{z_X}(p,q)\geq\delta_s$. Reversing the source order reverses the required channel order.
The margin $\delta_s$ specifies the minimum relative difference between the normalized channel responses.
For edge masks, $g_f$ is the two-component Sobel gradient and $s_f=\sqrt{\|g_f\|_2^2+\varepsilon}$.
The components of $g_f$ are horizontal and vertical derivatives. Their Euclidean norm is $\|g_f\|_2$, so $s_f$ measures edge strength rather than pixel brightness.
The horizontal kernel has rows $(-1,0,1)$, $(-2,0,2)$, and $(-1,0,1)$. The vertical kernel is its transpose, and both use zero padding of one pixel.
With spatial means $\bar s_I,\bar s_Y$ computed separately per image, define
\begin{equation}
\begin{aligned}
 c&=\frac{g_I^\top g_Y}{s_Is_Y+\varepsilon},\\
 M_s^{\rm edge}(p)&=\begin{cases}
 1,&\scriptstyle s_I(p)>\bar s_I,\ s_Y(p)>\bar s_Y,\ c(p)>\gamma_s,\\
 0,&\text{otherwise},
 \end{cases}\\
 M_m^{\rm edge}(p)&=\begin{cases}
 1,&\scriptstyle s_Y(p)>\bar s_Y,\
      \big(s_I(p)\leq\bar s_I\ \text{or}\ c(p)<\gamma_m\big),\\
 0,&\text{otherwise}.
 \end{cases}
\end{aligned}
\end{equation}
The numerator $g_I^\top g_Y$ is the inner product of the two gradient vectors at each pixel, and $s_Is_Y$ is the numerical product of their magnitudes.
The normalized alignment $c(p)$ is large and positive for similarly directed gradients.
The shared mask requires above-mean edges in both sources and alignment above $\gamma_s$. The candidate-material mask requires an above-mean visible edge together with either an infrared edge or alignment below $\gamma_m$.
Both masks are derived from the source images, with $\gamma_s$ and $\gamma_m$ controlling gradient alignment.
The $X$-edge loss preserves the former and suppresses the latter in the Texture map:
\begin{equation}
\begin{aligned}
    \mathcal{L}_{X\mathrm{edge}}
    &=\frac{\sum_p M_s^{\rm edge}(p)
       \left|s_{z_X}(p)-\tfrac12(s_I(p)+s_Y(p))\right|}
       {\max(1,\sum_p M_s^{\rm edge}(p))}\\
    &\quad+\lambda_{\rm edge}
       \frac{\sum_p M_m^{\rm edge}(p)s_{z_X}(p)^2}
       {\max(1,\sum_p M_m^{\rm edge}(p))},
\end{aligned}
\end{equation}
where $\lambda_{\rm edge}$ balances edge suppression.
Each fraction averages only over its own selected pixels, with $\sum_p$ including the current batch.
On shared edges, the first term matches $s_{z_X}$ to the mean source edge strength. On candidate material edges, the second term suppresses $s_{z_X}$ toward zero.
This suppression regularizes local channel gradients at candidate material boundaries.
The channel-structure weights $\lambda_s$, $\lambda_X$, and $\lambda_t$ balance the shading-order, $X$-edge, and temperature-order terms, respectively.

Thermal ordering uses a reliability mask to select infrared evidence for relative temperature guidance.
Let $\mu_r=\operatorname{AvgPool}_{k_r}(I)$ and $\nu_r=\operatorname{AvgPool}_{k_r}((I-\mu_r)^2)$ be local intensity statistics, with resolution-preserving pooling of width $k_r$.
Here, $\operatorname{AvgPool}_{k_r}$ computes a local mean, $\mu_r$ is the smoothed infrared image, and $\nu_r$ is the locally averaged squared residual from that image. The quantity $\sqrt{\nu_r}$ measures local activity.
The point and pair masks are
\begingroup
\setlength{\arraycolsep}{2pt}
\begin{equation}
\begin{aligned}
 r(p)&=\begin{cases}
 1,&\scriptstyle \sqrt{\nu_r(p)}>\tau_r,\
       \mathcal{D}_e(z_e(p))>\tau_e^{\rm rel},\ |I(p)-z_X(p)|<\tau_{IX},\\
 0,&\text{otherwise},
 \end{cases}\\
 M_t(p,q)&=\begin{cases}
 1,&\scalebox{0.92}{$\scriptstyle r(p)=r(q)=1,\ |I(p)-I(q)|>\tau_I,\ |z_e(p)-z_e(q)|<\tau_e^{\rm pair}$},\\
 0,&\text{otherwise}.
 \end{cases}
\end{aligned}
\end{equation}
\endgroup
The thresholds select local activity, sufficient decoded emissivity, a limited difference between $I$ and $z_X$, and pixel pairs with sufficient infrared contrast with similar values in the normalized Emissivity map.
Here, $\tau_r$ is the minimum local infrared activity, $\tau_e^{\rm rel}$ is the minimum decoded emissivity for the reliability mask, and $\tau_{IX}$ bounds the difference between normalized $I$ and $z_X$.
The pair threshold $\tau_I$ requires sufficient absolute infrared contrast, whereas $\tau_e^{\rm pair}$ limits the absolute difference between the normalized Emissivity codes at the two endpoints.
The binary mask $M_t$ records pair eligibility: both endpoints must pass $r$, and the pair must satisfy both pairwise tests.
Only pairs with $M_t=1$ contribute to the temperature-order term. Excluded pairs may still contribute to other losses.
The temperature-order loss uses margin $\delta_t$:
\begin{equation}
\begin{aligned}
    \mathcal{L}_{\rm temp}
    &=\frac{1}{\max(1,|\mathcal{O}_t|)}
      \sum_{o\in\mathcal{O}_t}\frac{1}{N_t^o}
      \sum_{(p,q)\in\mathcal{P}_o}M_t(p,q)\\
    &\qquad\cdot\max\left(0,\delta_t-d_{z_T}\operatorname{sign}(d_I)\right).
\end{aligned}
\end{equation}
Here, $N_t^o=N_{M_t}^o$ and $\mathcal{O}_t=\mathcal{O}_{M_t}$ preserve the same within-offset and across-offset averaging as the shading-order term.
For an eligible pair, zero penalty requires $d_{z_T}\operatorname{sign}(d_I)\geq\delta_t$: normalized Temperature must follow the infrared ordering with at least the relative margin $\delta_t\geq0$.

A contrast prior stabilizes the initial $T$ and $X$ responses.
For $\hat I=I/(\max_x I+\varepsilon)$ and $\hat Y=Y/(\max_x Y+\varepsilon)$, define
\begin{equation}
\begin{aligned}
 t_*&=\operatorname{sigmoid}(\kappa(\hat I-\hat Y)),\\
 x_*&=\operatorname{sigmoid}(\kappa(\operatorname{AvgPool}_{k_p}(\hat Y)-\hat I)),\\
 \mathcal{L}_{\rm prior}&=\lambda_{\rm contrast}\big(\operatorname{MSE}(z_T,t_*)\\
       &\qquad+\operatorname{MSE}(z_X,x_*)\big)+\lambda_{\rm range}P(z_T).
\end{aligned}
\end{equation}
Here, $\max_x$ takes the maximum over spatial positions of each image, the position index $x$ is unrelated to $X$ map.
The function $\operatorname{sigmoid}(u)=1/(1+\exp(-u))$ maps a scalar to $(0,1)$, $\kappa$ controls its response scale, and $k_p$ sets the local pooling width.
The soft target $t_*$ increases where normalized infrared intensity exceeds visible luminance, whereas $x_*$ increases where locally averaged visible luminance exceeds infrared intensity.
The $_*$ marks independently computed soft targets derived from the source images.
$\operatorname{MSE}$ denotes the mean squared error, while $\lambda_{\rm contrast}$ and $\lambda_{\rm range}$ balance the contrast and range-control terms.
Writing $\bar z_T$ for the spatial mean and $a(z_T)=a_-$ below that mean and $a_+$ otherwise, the range-control penalty is
\begin{equation}
\begin{aligned}
    P(z_T)&=\frac{1}{N}\sum_{p\in\Omega}
       \max\left(0,a(z_T(p))(z_T(p)-\bar z_T)^2-\delta_P\right)\\
       &\quad+\frac{\lambda_P}{N}\sum_{p\in\Omega}|z_T(p)-\bar z_T|,
\end{aligned}
\end{equation}
where $a_-$ and $a_+$ control asymmetric spread, $\delta_P$ is the spread tolerance, and $\lambda_P$ weights absolute deviation.
The mean $\bar z_T=N^{-1}\sum_{p\in\Omega}z_T(p)$ is computed from the current image.
The coefficient $a(z_T(p))$ weights deviations by $a_-$ below that mean and $a_+$ otherwise.
The first term penalizes squared deviations only above its tolerance, while the second penalizes absolute deviations throughout.
Thus, $P$ regularizes spatial spread around the current image mean, while the reconstruction and structural terms guide scene-dependent variation.
The per-image range penalties are averaged over the batch.
The channel-structure objective combines these terms with nonnegative weights:
\begin{equation}
    \mathcal{L}_{\rm ch}=\mathcal{L}_{\rm prior}
       +\lambda_s\mathcal{L}_{\rm shade}+\lambda_X\mathcal{L}_{X\mathrm{edge}}
       +\lambda_t\mathcal{L}_{\rm temp}.
\end{equation}

\subsubsection{Material Consistency}
\begin{figure}[!t]
    \centering
    \includegraphics[pagebox=cropbox,width=0.9\columnwidth]{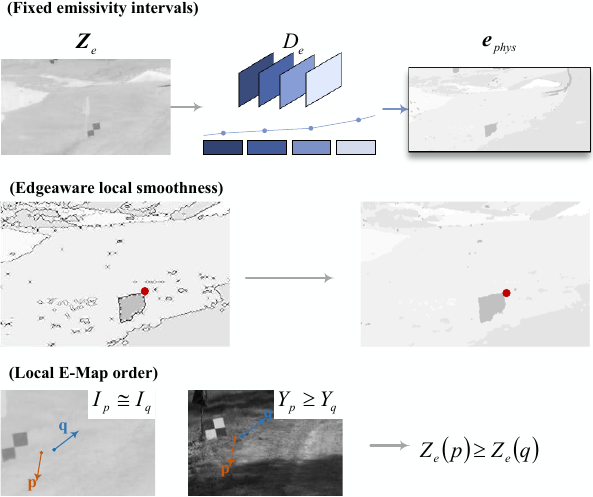}
    \caption{Material consistency through fixed-interval emissivity decoding, edge-aware local smoothness, and local ordering. The ordering schematic uses visible contrast to guide relative $z_e$ responses for infrared-similar pixel pairs.}
    \label{Fig_Method_TeXMetric_Weak_e}
\end{figure}

The emissivity decoder supplies a fixed interval prior, while local smoothness and ordinal guidance regularize the Emissivity map without material labels (Fig.~\ref{Fig_Method_TeXMetric_Weak_e}).
Define the detached edge $E_{\mathrm{edge}}=\max(s_I,s_Y)/\max(\varepsilon,\max_x\max(s_I,s_Y))$.
The inner maximum selects the stronger source edge at each pixel, and the spatial maximum normalizes it per image, giving $E_{\mathrm{edge}}\in[0,1]$.
This map is distinct from the reconstruction error $E_{\mathrm{rec}}$ and receives no gradient.
Edge-aware smoothness is
\begin{equation}
\begin{aligned}
    S(z_e)&=\sum_{j\in\{x,y\}}\frac{1}{|\Omega_j|}\sum_{p\in\Omega_j}
       |z_e(p+\mathbf{u}_j)-z_e(p)|\\
       &\qquad\cdot\exp\left(-\beta_E E_{\mathrm{edge}}(p+\mathbf{u}_j)\right),
\end{aligned}
\end{equation}
where $\mathbf{u}_j$ is a one-pixel horizontal or vertical offset, $\Omega_j$ contains the pixels whose neighbor $p+\mathbf{u}_j$ is valid, and $|\Omega_j|$ is their number.
The two directions are averaged separately before summation, and $\beta_E\geq0$ controls attenuation at edges.
The exponential provides continuous weights near one in flat regions and smaller near strong edges.
This weighting encourages smoothness within regions while allowing channel changes at source boundaries.
The material-order mask and loss are
\begin{equation}
\begin{aligned}
 M_e(p,q)&=\begin{cases}
 1,&\scriptstyle |I(p)-I(q)|<\tau_m,\ |d_Y|>\tau_s,\\
 0,&\text{otherwise},
 \end{cases}\\
 \mathcal{L}_{\rm material}
    &=\frac{1}{\max(1,|\mathcal{O}_e|)}
      \sum_{o\in\mathcal{O}_e}\frac{1}{N_e^o}
      \sum_{(p,q)\in\mathcal{P}_o}M_e(p,q)\\
    &\qquad\cdot\max\left(0,\delta_e-d_{z_e}\operatorname{sign}(d_Y)\right),\\
 \mathcal{L}_{\rm mat}&=\lambda_{\rm smooth}S(z_e)+\lambda_{\rm ordinal}\mathcal{L}_{\rm material}.
\end{aligned}
\end{equation}
Here, $N_e^o=N_{M_e}^o$ and $\mathcal{O}_e=\mathcal{O}_{M_e}$ count eligible pairs and nonempty offsets for material ordering.
The threshold $\tau_m$ selects infrared-similar pairs, $\delta_e\geq0$ is a relative Emissivity ordering margin, and $\lambda_{\rm smooth}$ and $\lambda_{\rm ordinal}$ weight edge-aware smoothness and material-order guidance, respectively.
The visible contrast test reuses the relative-difference threshold $\tau_s$, and both tests must hold for $M_e=1$.
For an eligible pair with $Y(p)>Y(q)$, the loss encourages $d_{z_e}(p,q)\geq\delta_e$, imposing a minimum relative difference between the normalized Emissivity responses.
Visible contrast provides local ordering guidance for $z_e$, while source-derived structural constraints regulate $z_X$. 
Their complementary roles are coupled through radiometric reconstruction.

\subsubsection{Complete Generation Objective}
The complete TeXMetric construction objective is
\begin{equation}
    \mathcal{L}_{\mathrm{con}}=\sum_{k\in\{\mathrm{rad,ch,mat}\}}\lambda_k\mathcal{L}_k.
\end{equation}
Here, $k$ indexes the radiometric reconstruction, channel-structure, and material loss groups, and the nonnegative weights $\lambda_k$ balance their contributions.
The objective updates the encoders, four feature-construction pathways, the three HyperConnection gates, and all three generation branches.
The $X$-structure control activates only $\mathcal{L}_{\rm shade}$ and $\mathcal{L}_{X\mathrm{edge}}$, and the material control activates only $\mathcal{L}_{\rm mat}$.
Generation validation assesses agreement with TeXMetric, while the generation objective contains no downstream task-loss term.

\subsection{TeX Tune}
\subsubsection{Frozen Generation and Trainable Tuning}
Let $G_{\theta^*}$ denote the generator selected after TeX generation, $\mathcal{A}_\phi$ the channel-adjustment module, and $D_\psi$ the complete downstream task network.
Every parameter of $G_{\theta^*}$, including the encoders, the three HyperConnection gates, and the generation branches, remains frozen.
The adjustment module and task network remain trainable.
For each visible-infrared image pair, the forward path is
\begin{equation}
\begin{aligned}
    Z^b&=\operatorname{sg}\left(G_{\theta^*}(I_v,I_{ir})\right),\\
    Z^a&=\mathcal{A}_\phi(Z^b),
    \hat{y}^{\,a}=D_\psi(Z^a),
\end{aligned}
\end{equation}
where $\operatorname{sg}$ stops gradients and the superscripts $b$ and $a$ denote the constructed baseline and task-tuned TeX, respectively.
The parameter sets $\theta^*$, $\phi$, and $\psi$ belong to the selected generator, tune module, and task network, respectively, and $\hat y^{\,a}$ is the downstream prediction from tuned TeX.
Multi-scale features are extracted directly from the generated TeX within $\mathcal{A}_\phi$.

\subsubsection{Channel-Wise Tuning}
Three parallel branches operate independently on $Z_T^b$, $Z_e^b$, and $Z_X^b$.
Their architectures are identical, their parameters are separate, and the formulation below describes any channel $k\in\{T,e,X\}$.

For each generated channel $Z_k^b$, Multi-scale Guide learns features at multiple spatial scales,
\begin{equation}
    B_{k,\ell}=\mathcal{M}_{k,\ell}(Z_k^b),\qquad \ell\in\mathcal{L},
\end{equation}
where $\mathcal{L}$ indexes the feature scales and $\mathcal{M}_{k,\ell}$ are trainable feature-extraction mappings.
The features are projected to a common channel dimension and resized to the TeX resolution to form
\begin{equation}
\begin{aligned}
    \tilde B_{k,\ell}&=\operatorname{Up}_{H,W}(P_{k,\ell}(B_{k,\ell})),\\
    C_k&=\operatorname{Concat}\left(P_k(Z_k^b),
               \tilde B_{k,1},\ldots,\tilde B_{k,L}\right),
\end{aligned}
\end{equation}
where $P_k$ and $P_{k,\ell}$ are learned projections, $\operatorname{Up}_{H,W}$ restores the image resolution, and the $L$ scales in $\mathcal{L}$ are indexed from one to $L$.
Fine-scale features retain local detail, while coarse-scale features capture broader spatial context.

Local Activity Gate measures local variation in the corresponding baseline channel.
With $a_k=|Z_k^b-\operatorname{AvgPool}(Z_k^b)|$, the gate is
\begin{equation}
    G_k^{\mathrm{loc}}=\operatorname{sigmoid}
       \left(\frac{a_k-\mu(a_k)}{s(a_k)+\varepsilon}\right),
\end{equation}
where $\mu$ and $s$ denote the spatial mean and standard deviation.
The gate emphasizes local transitions and suppresses homogeneous regions, while task relevance is learned by the region-aware branch.

The Region-Aware module uses $C_k$ to predict both a residual $\Delta Z_k=R_k(C_k)$ and a spatial gate $G_k^{\mathrm{reg}}=\operatorname{sigmoid}(A_k(C_k))$.
Their combination gives
\begin{equation}
\begin{aligned}
    G_k=G_k^{\mathrm{loc}}\odot G_k^{\mathrm{reg}},\qquad
    Z_k^a=Z_k^b+\eta_kG_k\odot\Delta Z_k.
\end{aligned}
\end{equation}
Here, $R_k$ predicts the channel residual, $A_k$ predicts the spatial-gate logits, $G_k$ is the product of the local-activity and region-aware gates, and $\eta_k$ is a learned residual scale, and the superscript $a$ denotes task tuning.
The element-wise product of the two continuous gates spatially modulates the channel residual.
Concatenating the three tuned channels gives $Z^a=\operatorname{Concat}(Z_T^a,Z_e^a,Z_X^a)$.
The multi-scale feature extractors, residual predictors, region gates, guidance projections, and residual scales are included in $\phi$ and optimized with downstream supervision.

\subsubsection{Downstream Supervision and Relative Feedback}
The complete task network consumes $Z^a$ after channel adjustment.
Its task loss updates both $\phi$ and $\psi$, allowing channel responses and their downstream interpretation to be optimized together.
During training, an additional evaluation of the unadjusted $Z^b$ supplies the baseline loss used in the relative penalty.
The relative penalty updates $\phi$ while holding the task-network parameters fixed for that gradient path.
Downstream metrics assess validation performance, while the corresponding differentiable task loss supplies the training signal.
This task-derived feedback is motivated by the TDHF finding that visual-quality constraints need not align with task accuracy~\cite{li2025tdhf}.
The task criterion therefore determines the direction of channel adjustment, while the frozen generator preserves the common TeX generation.

\subsubsection{Tuning Objective}
For task labels $y$, the tuned task loss and detached baseline are
\begin{equation}
\begin{aligned}
    \ell_a=\mathcal{L}_{\mathrm{task}}(D_\psi(Z^a),y),\quad
    \ell_b=\operatorname{sg}\big(\mathcal{L}_{\mathrm{task}}(D_\psi(Z^b),y)\big).
\end{aligned}
\end{equation}
The task loss updates both the adjustment module and the complete downstream network.
For task penalty, the task network is evaluated with its parameters held constant while retaining derivatives with respect to its input:
\begin{equation}
    \tilde\ell_a=\mathcal{L}_{\mathrm{task}}(D_{\operatorname{sg}(\psi)}(Z^a),y).
\end{equation}
Here $D_{\operatorname{sg}(\psi)}$ uses the same current task-network parameters as $D_\psi$, but blocks their gradients only in the penalty path.
The task penalty term is
\begin{equation}
    \mathcal{L}_{\mathrm{up}}=\max(0,\tilde\ell_a-\ell_b).
\end{equation}
The penalty is activated when the tuned task loss exceeds the detached baseline loss, discouraging adjustments that increase the current training loss.
The complete objective is
\begin{equation}
    \mathcal{L}_{\mathrm{tune}}=\ell_a+
       \lambda_{\mathrm{up}}\mathcal{L}_{\mathrm{up}}.
\end{equation}
The coefficient $\lambda_{\mathrm{up}}$ balances the task penalty loss and the tuned task loss.
During training, $\mathcal{L}_{\mathrm{up}}$ updates only $\phi$, whereas $\ell_a$ updates both $\phi$ and $\psi$. The generator parameters $\theta^*$ remain fixed throughout tuning.
At inference, the frozen generator and adjustment module produce $Z^a$ for prediction by $D_\psi$, without requiring task labels or the baseline comparison.

\begin{algorithm}[t]
\caption{TeX Generation and TeX Tune.}
\label{alg:tex_pipeline}
\begin{algorithmic}[1]
\STATE Initialize $G_\theta$, adjustment module $\mathcal{A}_\phi$, and complete task network $D_\psi$.
\STATE \textbf{TeX Generation:} Train $G_\theta$ with the complete TeXMetric constraints.
\STATE Select $\theta^*$ using the generation validation criterion.
\STATE Freeze all parameters of $G_{\theta^*}$.
\FOR{each training pair and task annotation}
\STATE \textbf{TeX Tune:} Construct $Z^b=G_{\theta^*}(I_v,I_{ir})$.
\STATE Extract multi-scale features from the generated TeX within $\mathcal{A}_\phi$.
\STATE Tune $Z^a=\mathcal{A}_\phi(Z^b)$ with three independent adjustment branches.
\STATE Forward $Z^a$ through $D_\psi$ and compute $\ell_a$.
\STATE Compute the detached baseline $\ell_b$ and the penalty $\mathcal{L}_{\mathrm{up}}$.
\STATE Update $\phi$ using $\nabla_\phi(\ell_a+\lambda_{\mathrm{up}}\mathcal{L}_{\mathrm{up}})$.
\STATE Update $\psi$ using $\nabla_\psi\ell_a$, keeping $\theta^*$ fixed.
\ENDFOR
\STATE \textbf{Inference:} Generate $Z^b$ with frozen $G_{\theta^*}$.
\STATE Tune $Z^b$ to $Z^a$ and predict with $D_\psi(Z^a)$.
\end{algorithmic}
\end{algorithm}

\section{Experiments}

\begin{table}[!t]
\centering
\caption{Detection results on FLIR (\%). N/A indicates a value not reported in the corresponding publication.}
\label{Tab_Experiment_FLIR_Comparison}
\scriptsize
\setlength{\tabcolsep}{3pt}
\begin{tabular*}{\columnwidth}{@{\extracolsep{\fill}}l|c|c|ccc}
\toprule Method & Year & Modality & $mAP_{50}$ & $mAP_{75}$ & $mAP_{50:95}$\\
\midrule
DDQ-DETR~\cite{zhang2023dense} & 2023 & RGB & 73.9 & 32.2 & 37.1\\
YOLOv5~\cite{yolov5} & 2020 & RGB & 67.8 & 25.9 & 31.8\\
DDQ-DETR~\cite{zhang2023dense} & 2023 & IR & 64.9 & 24.5 & 30.9\\
YOLOv5~\cite{yolov5} & 2020 & IR & 73.9 & 35.7 & 39.5\\
GAFF~\cite{zhang2021guided} & 2021 & IR+RGB & 74.6 & 31.3 & 37.4\\
ProbEn~\cite{chen2022multimodal} & 2022 & IR+RGB & 75.5 & 31.8 & 37.9\\
CFT~\cite{qingyun2022cross} & 2022 & IR+RGB & 78.0 & 35.5 & 40.2\\
TarDAL~\cite{liu2022target} & 2022 & IR+RGB & 79.9 & 37.9 & N/A\\
CSSA~\cite{cao2023multimodal} & 2023 & IR+RGB & 79.2 & 37.4 & 41.3\\
LRAF-Net~\cite{fu2023lraf} & 2024 & IR+RGB & 80.5 & N/A & 42.8\\
MMI-Det~\cite{zeng2024mmi} & 2024 & IR+RGB & 79.8 & 35.8 & 40.5\\
WaveMamba~\cite{zhu2025wavemamba} & 2025 & IR+RGB & 88.4 & N/A & 48.1\\
IC-Fusion~\cite{hwang2025multispectral} & 2025 & IR+RGB & 83.3 & 42.6 & 46.1\\
TDHF-YOLOv5~\cite{li2025tdhf} & 2026 & IR+RGB & 83.8 & 45.1 & 46.2\\
Distribution-Aligned~\cite{hao2026distribution} & 2026 & IR+RGB & 79.5 & N/A & N/A\\
FreDFT~\cite{wu2026fredft} & 2026 & IR+RGB & 83.5 & N/A & 42.6\\
RegisterBridgeMM~\cite{wang2026registerbridgemm} & 2026 & IR+RGB & \textbf{88.9} & N/A & 49.8\\
MVR-MoE~\cite{li2026mvrmoe} & 2026 & IR+RGB & 85.7 & 53.2 & 51.8\\
\textbf{Ours} & 2026 & IR+RGB & 87.5 & \textbf{54.2} & \textbf{52.9}\\
\bottomrule
\end{tabular*}
\end{table}

\begin{table}[!t]
\centering
\caption{Detection results on LLVIP (\%). N/A indicates a value not reported in the corresponding publication.}
\label{Tab_Experiment_LLVIP_Comparison}
\scriptsize
\setlength{\tabcolsep}{3pt}
\begin{tabular*}{\columnwidth}{@{\extracolsep{\fill}}l|c|c|ccc}
\toprule Method & Year & Modality & $mAP_{50}$ & $mAP_{75}$ & $mAP_{50:95}$\\
\midrule
DDQ-DETR~\cite{zhang2023dense} & 2023 & RGB & 86.1 & 45.8 & 46.7\\
YOLOv5~\cite{yolov5} & 2020 & RGB & 90.8 & 51.9 & 50.0\\
DDQ-DETR~\cite{zhang2023dense} & 2023 & IR & 93.9 & 64.6 & 58.6\\
YOLOv5~\cite{yolov5} & 2020 & IR & 94.6 & 72.2 & 61.9\\
GAFF~\cite{zhang2021guided} & 2021 & IR+RGB & 94.0 & 60.2 & 55.8\\
ProbEn~\cite{chen2022multimodal} & 2022 & IR+RGB & 93.4 & 50.2 & 51.5\\
CFT~\cite{qingyun2022cross} & 2022 & IR+RGB & 97.5 & 72.9 & 63.6\\
TarDAL~\cite{liu2022target} & 2022 & IR+RGB & 93.3 & 62.4 & N/A\\
CSSA~\cite{cao2023multimodal} & 2023 & IR+RGB & 94.3 & 66.6 & 59.2\\
LRAF-Net~\cite{fu2023lraf} & 2024 & IR+RGB & 97.9 & N/A & 66.3\\
MMI-Det~\cite{zeng2024mmi} & 2024 & IR+RGB & \textbf{98.9} & 73.5 & 64.4\\
GAN-HA~\cite{lu2024gan} & 2024 & IR+RGB & 94.3 & N/A & 56.0\\
WaveMamba~\cite{zhu2025wavemamba} & 2025 & IR+RGB & 98.3 & N/A & 66.0\\
IC-Fusion~\cite{hwang2025multispectral} & 2025 & IR+RGB & 97.9 & 81.5 & 70.3\\
TDHF-YOLOv5~\cite{li2025tdhf} & 2026 & IR+RGB & 97.5 & 78.3 & 67.4\\
Distribution-Aligned~\cite{hao2026distribution} & 2026 & IR+RGB & 98.1 & N/A & N/A\\
FreDFT~\cite{wu2026fredft} & 2026 & IR+RGB & 98.2 & N/A & 66.2\\
RegisterBridgeMM~\cite{wang2026registerbridgemm} & 2026 & IR+RGB & 98.4 & N/A & 70.5\\
InfraNet~\cite{feng2026infranet} & 2026 & IR+RGB & 97.2 & N/A & 67.6\\
MVR-MoE~\cite{li2026mvrmoe} & 2026 & IR+RGB & 98.2 & 84.0 & 72.0\\
\textbf{Ours} & 2026 & IR+RGB & 97.9 & \textbf{85.2} & \textbf{72.7}\\
\bottomrule
\end{tabular*}
\end{table}

\begin{table}[t]
\centering
\caption{Aggregate detection results on M3FD (\%). N/A indicates a value not reported in the corresponding publication.}
\label{Tab_Experiment_M3FD_Aggregate_Comparison}
\scriptsize
\setlength{\tabcolsep}{3pt}
\begin{tabular*}{\columnwidth}{@{\extracolsep{\fill}}lcccc}
\toprule
Method & Year & Modality & $mAP_{50}$ & $mAP_{50:95}$\\
\midrule
CFT~\cite{qingyun2022cross} & 2022 & IR+RGB & 89.8 & 60.7\\
TarDAL~\cite{liu2022target} & 2022 & IR+RGB & 86.0 & 56.0\\
MetaFusion~\cite{zhao2023metafusion} & 2023 & IR+RGB & 86.7 & 56.8\\
CDDFuse~\cite{zhao2023cddfuse} & 2023 & IR+RGB & 87.0 & 56.5\\
ICAFusion~\cite{shen2024icafusion} & 2024 & IR+RGB & 87.4 & 57.2\\
Text-IF~\cite{yi2024text} & 2024 & IR+RGB & N/A & 40.4\\
CDTFusion~\cite{zhao2025cdtfusion} & 2025 & IR+RGB & 76.6 & 50.1\\
MTG-Fusion~\cite{wang2025multi} & 2025 & IR+RGB & N/A & 52.6\\
CoDAF~\cite{zongzhen2025cross} & 2025 & IR+RGB & \textbf{90.8} & 61.2\\
FreeFusion~\cite{zhao2025freefusion} & 2025 & IR+RGB & N/A & 62.6\\
FreDFT~\cite{wu2026fredft} & 2026 & IR+RGB & 88.4 & 59.7\\
InfraNet~\cite{feng2026infranet} & 2026 & IR+RGB & 89.9 & 64.0\\
MVR-MoE~\cite{li2026mvrmoe} & 2026 & IR+RGB & 88.8 & 63.4\\
\textbf{Ours} & 2026 & IR+RGB & 88.8 & \textbf{64.2}\\
\bottomrule
\end{tabular*}
\end{table}

\subsection{Datasets and Evaluation}
We evaluate TeX-Gen on visual tasks using object detection and semantic segmentation.
Detection is evaluated on FLIR~\cite{flir2021}, LLVIP~\cite{jia2021llvip}, and M3FD~\cite{liu2022target}. A separate comparison summarizes published nine-class semantic-segmentation results on MSRS~\cite{liu2023multi}.
FLIR contains aligned visible-thermal driving scenes with four object categories, LLVIP focuses on low-light pedestrian detection, and M3FD contains six categories across traffic and surveillance scenes.
We report $mAP_{50}$, $mAP_{75}$, and $mAP_{50:95}$ for detection. The MSRS comparison lists published class-wise intersection over union (IoU) and mean IoU ($mIoU$) over nine classes, including Unlabelled (Background).
The HADAR dataset~\cite{bao2023heat} is analyzed separately using its available reference channels.
The same channel encoding is used for downstream task inputs and channel visualization across all compared generation objectives.
For radiometric reconstruction, $X$ is computed from $V$ through the environmental-radiation relation defined in the method section~\cite{bao2023heat}. 
Comparing $V$ aligns the experimental representation with the HADAR channels, while $X$ enters the radiometric reconstruction term~\cite{li2026mvrmoe}.
The radiometric formulation uses $T_{\min}=275$ K and $T_{\max}=325$ K, with fixed environmental temperatures $T_1=293.15$ K and $T_2=298.15$ K.
The training schedules use 200 TeX Generation epochs followed by 100 TeX Tune epochs.

\subsection{Comparison with Existing Methods}
\subsubsection{FLIR Results}
Table~\ref{Tab_Experiment_FLIR_Comparison} compares single-sensor and paired-sensor methods on FLIR.
The proposed method obtains the highest reported $mAP_{75}$ and $mAP_{50:95}$, exceeding MVR-MoE by 1.0 and 1.1 percentage points, respectively.
The result demonstrates improved localization performance over MVR-MoE at stricter thresholds.

\subsubsection{LLVIP Results}
Table~\ref{Tab_Experiment_LLVIP_Comparison} reports the LLVIP comparison.
The stricter localization scores exceed the MVR-MoE results by 1.2 and 0.7 percentage points, respectively.
The mAP50:95 score also exceeds the reported RegisterBridgeMM, InfraNet, and IC-Fusion results.
These results show improved localization at higher IoU thresholds on LLVIP.

\subsubsection{M3FD Results}
\begin{figure}[t]
    \centering
    \includegraphics[width=0.9\columnwidth]{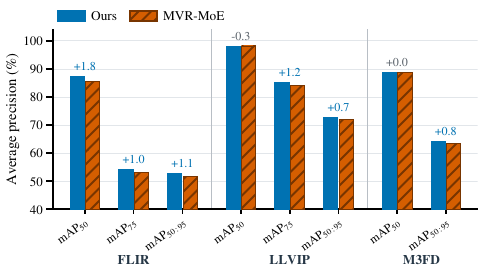}
    \caption{Detection comparison summarized from Tables~\ref{Tab_Experiment_FLIR_Comparison} through~\ref{Tab_Experiment_M3FD_Aggregate_Comparison}. Solid blue and hatched orange bars denote Ours and MVR-MoE, respectively, and $\Delta$ is Ours minus MVR-MoE in AP percentage points.}
    \label{Fig_Experiment_Detection_Summary}
\end{figure}

Table~\ref{Tab_Experiment_M3FD_Aggregate_Comparison} compares methods on M3FD using $mAP_{50}$ and $mAP_{50:95}$.
The proposed method matches MVR-MoE in $mAP_{50}$ and exceeds it by 0.8 percentage points in $mAP_{50:95}$.
Its $mAP_{50:95}$ is the highest among the listed results, while CoDAF reports the highest $mAP_{50}$.
These results show that TeX-Gen maintains competitive performance on M3FD.
Fig.~\ref{Fig_Experiment_Detection_Summary} summarizes the comparison with MVR-MoE across the three detection datasets.


\subsubsection{MSRS Semantic Segmentation}
Table~\ref{Tab_Experiment_MSRS_9Class} compares published MSRS results with our evaluation over nine semantic classes, including Unlabelled (Background), using class-wise IoU and nine-class $mIoU$.
Among the listed methods evaluated over all nine classes, our method achieves the highest $mIoU$ of 77.33\%, exceeding TDFusion by 2.24 percentage points.
TeX-Gen also achieves higher IoU than MVR-MoE in all nine classes.

\begin{table*}[!t]
\centering
\caption{Nine-class MSRS semantic-segmentation results (\%). Unlabelled denotes Background, and $mIoU$ includes all nine classes.}
\label{Tab_Experiment_MSRS_9Class}
\scriptsize
\setlength{\tabcolsep}{2.5pt}
\begin{tabular*}{\textwidth}{@{\extracolsep{\fill}}l|c|c|ccccccccc|c}
\toprule
Method & Year & Modality & Unlabelled & Car & Person & Bike & Curve & Car Stop & Guardrail & Color Cone & Bump & $mIoU$ \\
\midrule
EMMA~\cite{zhao2024emma} & 2024 & IR+RGB & 98.37 & 89.32 & 71.15 & 70.58 & 60.71 & 70.37 & 69.59 & 64.75 & 75.50 & 74.48 \\
DCINN~\cite{wang2024dcinn} & 2024 & IR+RGB & 98.34 & 89.08 & 70.94 & 69.19 & 59.31 & 73.27 & 72.21 & 64.02 & 72.76 & 74.35 \\
MRFS~\cite{zhang2024mrfs} & 2024 & IR+RGB & 98.32 & 88.80 & 70.55 & 70.41 & 58.76 & \textbf{73.85} & 74.66 & 62.80 & 72.37 & 74.50 \\
TIMFusion~\cite{liu2024timfusion} & 2024 & IR+RGB & 98.30 & 89.19 & 68.68 & 69.64 & 59.55 & 71.94 & 69.48 & 63.69 & 71.75 & 73.58 \\
TDFusion~\cite{tdfusion2025} & 2025 & IR+RGB & \textbf{98.38} & \textbf{89.75} & \textbf{71.76} & 70.36 & 59.77 & 72.40 & 75.06 & 64.61 & 73.73 & 75.09 \\
KDFuse~\cite{yang2025kdfuse} & 2025 & IR+RGB & 98.20 & 87.50 & 69.70 & 60.50 & 64.70 & 60.60 & 85.50 & 46.90 & 82.90 & 73.00 \\
GIFNet~\cite{cheng2025gifnet} & 2025 & IR+RGB & 98.10 & 87.20 & 69.30 & 60.70 & 61.50 & 59.10 & 79.20 & 46.60 & 81.60 & 71.50 \\
DAFusion~\cite{wang2025dafusion} & 2025 & IR+RGB & 98.00 & 87.30 & 69.30 & 59.50 & 58.90 & 59.90 & 76.20 & 44.80 & 77.30 & 70.10 \\
SAGE~\cite{wu2025sage} & 2025 & IR+RGB & 98.00 & 87.10 & 68.40 & 59.60 & 60.00 & 55.70 & 85.10 & 45.50 & 80.50 & 71.10 \\
FusionBooster~\cite{cheng2025fusionbooster} & 2025 & IR+RGB & 98.10 & 86.90 & 70.20 & 61.40 & 62.10 & 60.30 & 82.90 & 45.80 & 81.80 & 72.20 \\
MISSFusion~\cite{he2026missfusion} & 2026 & IR+RGB & 98.20 & 87.70 & 69.80 & 61.60 & 65.70 & 61.00 & 84.90 & 46.70 & \textbf{83.60} & 73.20 \\
MVR-MoE~\cite{li2026mvrmoe} & 2026 & IR+RGB & 98.05 & 83.37 & 70.64 & 69.44 & 61.94 & 68.87 & 86.36 & 54.64 & 64.47 & 73.09 \\
\textbf{Ours} & 2026 & IR+RGB & 98.28 & 85.23 & 71.43 & \textbf{71.76} & \textbf{72.68} & 72.18 & \textbf{86.64} & \textbf{65.60} & 72.19 & \textbf{77.33} \\
\bottomrule
\end{tabular*}
\end{table*}

\subsection{Ablation Studies}
\label{Sec_Experiment_Ablation}
\subsubsection{TeX Tune Components}

Table~\ref{Tab_Experiment_FLIR_Component_Ablation} evaluates the four TeX Tune components on FLIR. 
Each ablation removes one component while retaining the others under the same training and evaluation settings. 
The complete model outperforms each single-component ablation on the aggregate and scale-specific metrics shown here.
\begin{table}[t]
\centering
\caption{Ablation of TeX Tune components on FLIR.}
\label{Tab_Experiment_FLIR_Component_Ablation}
\scriptsize
\setlength{\tabcolsep}{2.2pt}
\renewcommand{\arraystretch}{1.12}
\renewcommand{\tabularxcolumn}[1]{m{#1}}
\begin{tabularx}{\columnwidth}{>{\raggedright\arraybackslash}X|*{6}{c}}
\toprule
Variant & $mAP_{50}$ & $mAP_{75}$ & $mAP_{50:95}$ & $mAP_s$ & $mAP_m$ & $mAP_l$ \\
\midrule
Full TeX Tune & \textbf{87.5} & \textbf{54.2} & \textbf{52.9} & \textbf{39.1} & \textbf{59.0} & \textbf{73.3}\\
\tableguideline
\textit{w/o} penalty term & 86.0 & 49.0 & 50.0 & 36.4 & 56.0 & 71.6\\
\tableguideline
\textit{w/o} multi-scale guide & 86.2 & 49.0 & 49.8 & 36.5 & 55.5 & 70.9\\
\tableguideline
\textit{w/o} local activity & 84.7 & 48.9 & 49.4 & 36.3 & 55.0 & 70.1\\
\tableguideline
\textit{w/o} region-aware branch & 85.3 & 49.3 & 50.0 & 36.7 & 55.6 & 71.7\\
\bottomrule
\end{tabularx}
\end{table}

\begin{table}[!tb]
\centering
\caption{Ablation of TeXMetric constraint groups on FLIR.}
\label{Tab_Experiment_Metric_Ablation}
\scriptsize
\setlength{\tabcolsep}{2.2pt}
\renewcommand{\arraystretch}{1.12}
\renewcommand{\tabularxcolumn}[1]{m{#1}}
\begin{tabularx}{\columnwidth}{>{\raggedright\arraybackslash}X|*{6}{c}}
\toprule
Generation criterion & $mAP_{50}$ & $mAP_{75}$ & $mAP_{50:95}$ & $mAP_s$ & $mAP_m$ & $mAP_l$\\
\midrule
Radiometric reconstruction & 80.5 & 45.4 & 46.2 & 32.1 & 52.5 & 70.9\\
\tableguideline
Reconstruction + $X$ structure & 81.8 & 46.0 & 47.3 & 33.5 & 53.9 & 68.3\\
\tableguideline
Reconstruction + material consistency & 81.9 & 45.1 & 47.0 & 33.3 & 53.2 & 69.4\\
\tableguideline
Full TeXMetric & \textbf{87.5} & \textbf{54.2} & \textbf{52.9} & \textbf{39.1} & \textbf{59.0} & \textbf{73.3}\\
\bottomrule
\end{tabularx}
\end{table}

\subsubsection{TeXMetric Constraint Contributions}
\begin{figure}[!t]
    \centering
    \includegraphics[width=0.95\columnwidth]{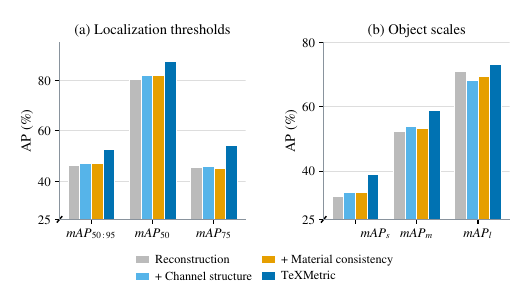}
    \caption{Downstream responses to TeXMetric constraints on FLIR, summarized from Table~\ref{Tab_Experiment_Metric_Ablation}.}
    \label{Fig_Experiment_Constraint_Comparison}
\end{figure}

Table~\ref{Tab_Experiment_Metric_Ablation} presents an ablation of TeXMetric constraint groups on FLIR.
The configurations retain radiometric reconstruction while evaluating channel-structure and material consistency terms individually and in the complete objective.
The complete TeXMetric objective achieves the highest performance across all six reported measures, as shown in Fig.~\ref{Fig_Experiment_Constraint_Comparison}.

\subsection{TeX Tune Analysis}
We evaluate task-guided adjustment on FLIR by comparing direct-input TeX with task-adjusted TeX. 
We also compare joint optimization of the channel-adjustment modules and task network with adjustment-only optimization under a frozen task network. 
Progress is measured in TeX Tune epochs.

Both conditions use a 100-epoch TeX Tune. 
The adjustment-only setting freezes the detector and updates only the channel-correction modules, whereas joint optimization also updates the detector feature extractor and prediction head.

\begin{table}[t]
\centering
\caption{FLIR input-configuration comparison and joint task tuning.}
\label{Tab_Experiment_FLIR_Optimization}
\scriptsize
\setlength{\tabcolsep}{3pt}
\renewcommand{\arraystretch}{1.12}
\begin{tabularx}{\columnwidth}{>{\raggedright\arraybackslash}X|ccc}
\toprule
\multicolumn{4}{c}{Direct-input control and task-adjusted TeX} \\
Condition & $mAP_{50}$ & $mAP_{75}$ & $mAP_{50:95}$ \\
\midrule
Generated TeX (without adjustment) & 85.5 & 47.8 & 49.3 \\
Task-adjusted TeX & \textbf{87.5} & \textbf{54.2} & \textbf{52.9} \\
\midrule
\multicolumn{4}{c}{Joint tuning ($mAP_{50:95}$)} \\
Trainable components & Best epoch & \multicolumn{2}{c}{\vphantom{\makecell{Best before\\end}}Best epoch before final epoch} \\
\midrule
Channel adjustment only (detector frozen) & 100 & \multicolumn{2}{c}{No} \\
Channel adjustment and detector & 52 & \multicolumn{2}{c}{Yes} \\
\bottomrule
\end{tabularx}
\end{table}

\begin{figure}[!t]
    \centering
    \includegraphics[width=0.9\columnwidth]{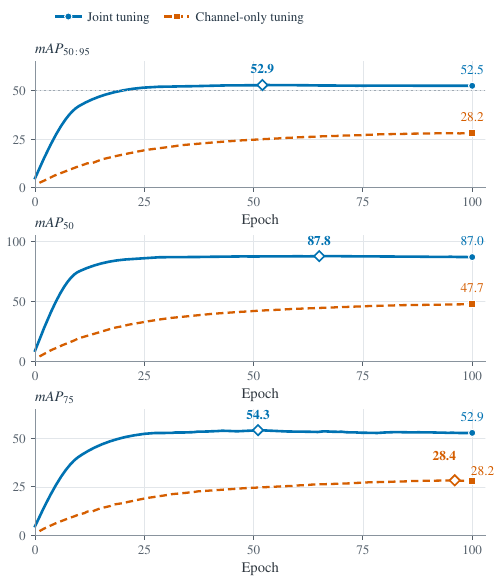}
    \caption{Joint task tuning on FLIR dataset during TeX Tune. Updating channel corrections alone with the inherited detector frozen makes limited progress, whereas jointly updating the corrections and task branch enables the task network to learn from the tuned TeX.}
    \label{Fig_Experiment_FLIR_Optimization_Trajectory}
\end{figure}

As shown in the upper block of Table~\ref{Tab_Experiment_FLIR_Optimization} , task-adjusted TeX improves $mAP_{50}$, $mAP_{75}$, and $mAP_{50:95}$ by 2.0, 6.4, and 3.6 percentage points, respectively, over direct-input TeX.
Joint optimization reaches its best recorded $mAP_{50:95}$ at epoch 52, whereas channel-only optimization reaches its best value at the last epoch, as shown in Fig.~\ref{Fig_Experiment_FLIR_Optimization_Trajectory}.
The results indicate that joint optimization benefits from updating both the channel responses and their downstream processing.

\begin{figure}[!t]
    \centering 
    \includegraphics[width=0.9\columnwidth]{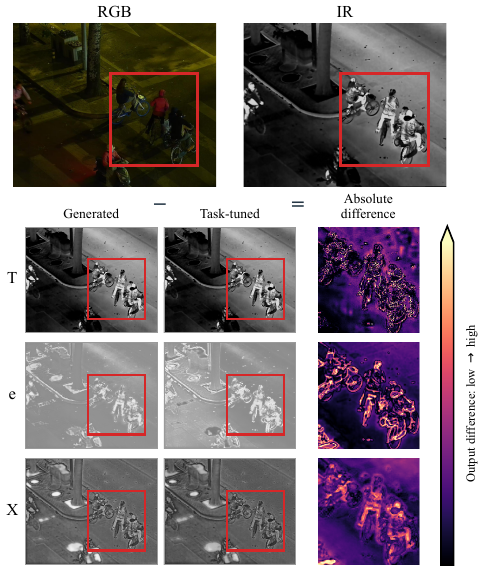}
\caption{Local TeX changes after task tuning on LLVIP. Brighter pixels in the cropped absolute-difference maps indicate larger changes.}
    \label{Fig_Experiment_TeX_Adaptation}
\end{figure}

\subsection{Visualization Analysis}
Fig.~\ref{Fig_Experiment_TeX_Adaptation} compares TeX from the best TeX Generation checkpoint with task-tuned TeX from the best TeX Tune checkpoint and visualizes their absolute differences.
The comparison shows how the generated channels change after task-guided tuning.
The multi-pedestrian scene illustrates local changes around pedestrians, with the overall scene structure preserved.
Spatially varying corrections adjust local contrast and boundary responses within the generated scene representation.

\section{Discussion}

\subsection{Connection Between TeXMetric and Downstream Task}
\label{Sec_Discussion_TeXMetric_Trajectory}
\begin{figure*}[!t]
    \centering
    \includegraphics[width=0.9\textwidth]{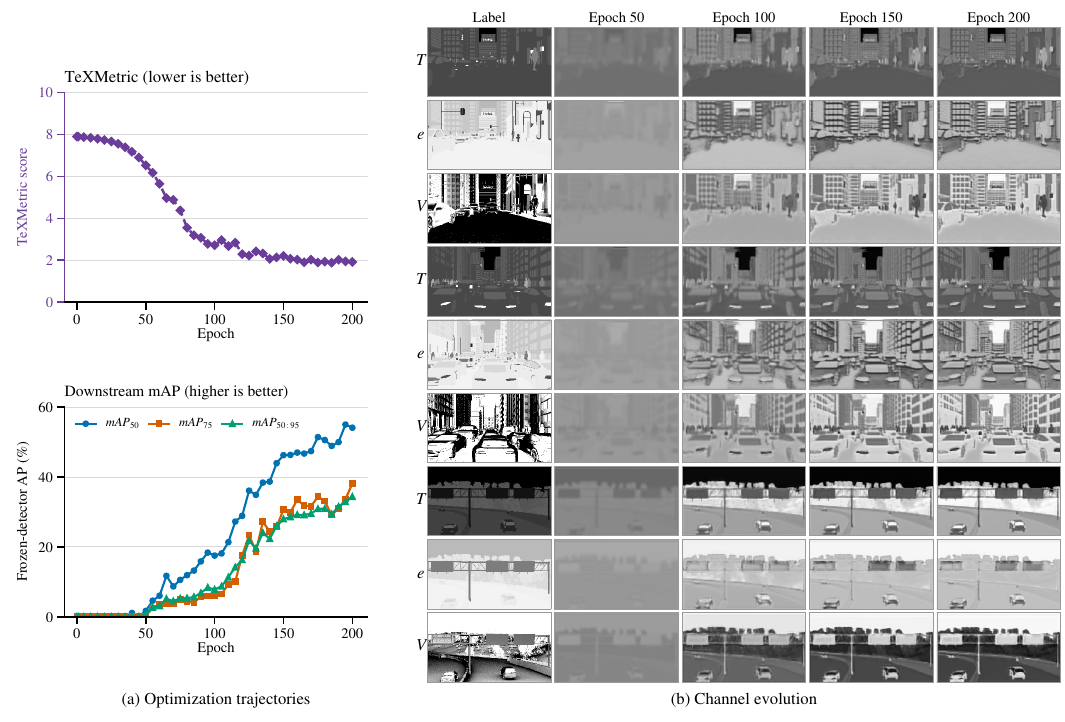}
\caption{TeXMetric improvement and downstream detection on HADAR.
    (a) Separate plots of TeXMetric score (lower is better) and AP under YOLOv8-S detector (higher is better), evaluated at the same generator checkpoints.
    (b) Ground-truth labels and generated maps at epochs 50, 100, 150, and 200 for street, crowded-street, and highway views.}
    \label{Fig_Experiment_HADAR_TeXMetric_Trajectory}
\end{figure*}
TeXMetric can also bring improvements to downstream tasks, which means that the accuracy of TeX in downstream downstream task performance improves as the TeXMetric score decreases.
As the generation process progressed, the complete TeXMetric score decreased from 7.90 to 1.92, while the detection performance (measured as $mAP_{50:95}$) improved to 34.4\%.
The relationship on the HADAR~\cite{bao2023heat} dataset is shown in Figure \ref{Fig_Experiment_HADAR_TeXMetric_Trajectory}.
This evaluation is based on 42 generator checkpoints and was conducted using images containing vehicles.
Furthermore, the generation process guided by TeXMetric is designed to preserve scene structure and object feature cues for low-light visual tasks.
As shown in Figure~\ref{Fig_Experiment_HADAR_TeXMetric_Trajectory}(b), the initially near-uniform T, e, and V responses gradually show clearer scene structure, boundaries of roads and buildings, and the structures of vehicles.

\subsection{Conventional Image-Quality Metrics vs. TeXMetric}
\begin{table}[!t]
\centering
\caption{FLIR detection results for radiometric reconstruction with conventional image-quality constraints.}
\label{Tab_Discussion_ImageQuality}
\scriptsize
\setlength{\tabcolsep}{3pt}
\begin{tabularx}{\columnwidth}{>{\raggedright\arraybackslash}X|ccc}
\toprule
Generation objective & $mAP_{50}$ & $mAP_{75}$ & $mAP_{50:95}$\\
\midrule
Radiometric reconstruction & 80.5 & 45.4 & 46.2\\
Radiometric reconstruction + Entropy & 81.4 & 43.8 & 45.7\\
Radiometric reconstruction + Gradient energy & 80.9 & 43.3 & 45.4\\
Radiometric reconstruction + MI & 81.9 & 44.6 & 46.2\\
Radiometric reconstruction + SSIM & 83.9 & 44.8 & 47.1\\
Radiometric reconstruction + PSNR & 82.5 & 43.5 & 45.9\\
Radiometric reconstruction + all five metrics & 82.7 & 44.8 & 46.8\\
\midrule
TeXMetric & \textbf{87.5} & \textbf{54.2} & \textbf{52.9}\\
\bottomrule
\end{tabularx}
\end{table}
Conventional image-quality metrics assess image appearance through information content, local variation, or similarity to RGB-IR images. 
However, these metrics are not suitable for TeX, as TeX is a visual representation of physical modalities. 
Consequently, we developed TeXMetric to evaluate TeX. 
It evaluates the radiometric, structural, and material relationships among the $T$, $e$, and $X$ maps.
We designed a comparative experiment between conventional metrics and TeXMetric to demonstrate that conventional image-quality metrics merely improve the recognizability of TeX as an image, without enhancing the accuracy of visual perception tasks involving TeX in low-light conditions.

On the FLIR dataset, Table~\ref{Tab_Discussion_ImageQuality} compares detection performance under four settings:
(1) radiometric reconstruction alone;
(2) its combination with each conventional image-quality metric (weight ratio of 0.5:0.5);
(3) its combination with all five metrics using weights of 0.5:0.1:0.1:0.1:0.1:0.1;
and (4) the three components of TeXMetric are assigned the same weights as those specified in the Experiments section.
For the conventional metrics, the generated grayscale $T$, $e$, and $X$ maps serve as the image-quality representations.
MI, SSIM, and PSNR additionally evaluate their similarity to the aligned visible and infrared input images.
The best results across the individual metric combinations for $mAP_{50}$, $mAP_{75}$, and $mAP_{50:95}$ are 83.9\%, 44.8\%, and 47.1\%, respectively.
When all five metrics are combined, the corresponding results are 82.7\%, 44.8\%, and 46.8\%, respectively.
In comparison, the TeXMetric objective achieves detection results of 87.5\%, 54.2\%, and 52.9\% on the three metrics, respectively.
The $mAP_{50:95}$ scores for these five combinations of individual metrics ranged from 45.4\% to 47.1\%.
Compared to the best-performing combination of traditional metrics, TeXMetric improved the $mAP_{50:95}$ score by 5.8 percentage points, while also achieving higher values for both $mAP_{50}$ and $mAP_{75}$.
These results indicate that, whether used individually or in combination, conventional image-quality metrics are less effective than TeXMetric in guiding TeX generation to optimize downstream detection performance.

\begin{table*}[!t]
\centering
\caption{Pixel-level supervision versus TeXMetric on HADAR under matched training budgets.}
\label{Tab_Discussion_HADAR_Detection}
\begin{tabular*}{\textwidth}{@{\extracolsep{\fill}}l|cccccc}
\toprule
TeX generation objective & MSE $\downarrow$ & RMSE $\downarrow$ & $mAP_{50}$ & $mAP_{75}$ & $mAP_{50:95}$ & $AR_{100}$\\
\midrule
Radiometric reconstruction & 0.1357 & 0.3684 & 50.4 & 26.5 & 28.7 & 40.2\\
Radiometric reconstruction + pixel MSE & 0.0555 & 0.2355 & 54.0 & 25.0 & 27.4 & 41.8\\
Radiometric reconstruction + pixel RMSE & \textbf{0.0535} & \textbf{0.2313} & 42.4 & 16.9 & 21.5 & 36.4\\
TeXMetric & 0.1319 & 0.3632 & \textbf{54.1} & \textbf{38.1} & \textbf{34.4} & \textbf{43.4}\\
\bottomrule
\end{tabular*}
\end{table*}

\subsection{Pixel-Level Supervision vs. TeXMetric Supervision}
\begin{figure}[!t]
    \centering
    \includegraphics[width=\columnwidth]{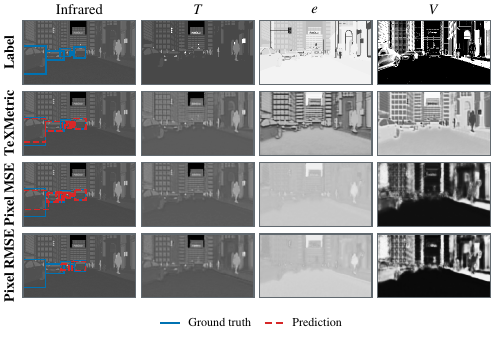}
    \caption{Downstream predictions and map results on HADAR for TeX generated under TeXMetric constraints versus pixel-level supervision (MSE and RMSE).}
    \label{Fig_Discussion_HADAR_Comparison}
\end{figure}

Pixel-level supervision directly minimizes the numerical differences between the generated channels and their labels, whereas TeXMetric constrains the radiometric, structural, and material relationships among the generated channels. 
HADAR provides normalized label maps of \(T\), \(e\), and \(V\) for this comparison, with \(X\) obtained from \(V\) through the affine radiometric mapping. 
All generation objectives retain radiometric reconstruction and use the same generator architecture, initialization, optimizer, and data augmentation strategy, differing only in the constraints applied. 
Both the generator and detector are trained on the training split and evaluated on the held-out test set. 
During evaluation, car, truck, and bus are merged into a single vehicle category.
Table~\ref{Tab_Discussion_HADAR_Detection} reveals a clear distinction between label fidelity and downstream detection performance. 
The pixel-supervised MSE and RMSE objectives yield lower label errors with respect to the labels, whereas TeXMetric yields larger errors. 
However, TeXMetric achieves the highest detection performance, with an $mAP_{50:95}$ of 34.4\%, compared with 27.4\% for pixel-level MSE and 21.5\% for pixel-level RMSE. 
This performance gap becomes more pronounced at stricter localization thresholds than at $mAP_{50}$.
Fig.~\ref{Fig_Discussion_HADAR_Comparison} presents the generated $T$, $e$, and $V$ maps together with their corresponding detection results. 
Blue solid boxes denote ground-truth annotations, while red dashed boxes denote predictions. 

The comparison separates label fidelity from downstream task performance. 
Under the evaluated HADAR training settings, pixel-supervised objectives produce lower label errors, whereas TeXMetric achieves higher detection AP. 
Thus, closer numerical agreement with the available channel labels does not necessarily translate into better downstream detection. 
This result is specific to the tested objectives and settings and does not establish that pixel-level supervision is generally inferior.

\section{Conclusion}

In this work, we presented TeX-Gen, which generates physically guided TeX maps from conventional visible and infrared images, reducing the reliance on specialized spectral imaging hardware. 
The proposed method organizes complementary information from infrared and visible images into Temperature, Emissivity, and Texture maps through online feature construction and channel-specific routing, providing a common TeX interface for downstream visual tasks. 
TeXMetric establishes a physically motivated criterion for TeX evaluation by combining radiometric reconstruction, channel structure, and material consistency without requiring pixel-level physical labels. 
Building on this common generation mapping, TeX Tune introduces task-guided channel corrections that improve downstream task performance while keeping the generator frozen. 
The HADAR comparison further reveals that closer agreement with pixel-level labels does not necessarily translate into better detection performance, highlighting the distinction between physical consistency and downstream task performance. 
These findings demonstrate the value of combining physically motivated channel generation with task-guided refinement, rather than relying on either numerical agreement or task performance alone to assess the generated channels.

\section*{Acknowledgments}
This work was funded by the Basic Research Program of Jiangsu under Grant No. BK20253028 and the National Natural Science Foundation of China under Grant Nos. U24A20330 and 62361166670.

\bibliographystyle{IEEEtran}
\bibliography{refernce}

@String(ICIP = {IEEE Int. Conf. Image Process.})

@String(AAAI = {AAAI})

@String(ICIP  = {ICIP})

@article{bayoudh2022survey,
  title={A survey on deep multimodal learning for computer vision: advances, trends, applications, and datasets},
  author={Bayoudh, Khaled and Knani, Raja and Hamdaoui, Fay{\c{c}}al and Mtibaa, Abdellatif},
  journal={The Visual Computer},
  volume={38},
  number={8},
  pages={2939--2970},
  year={2022},
  publisher={Springer}
}

@article{zhang2023visible,
  title={Visible and infrared image fusion using deep learning},
  author={Zhang, Xingchen and Demiris, Yiannis},
  journal={IEEE Transactions on Pattern Analysis and Machine Intelligence},
  volume={45},
  number={8},
  pages={10535--10554},
  year={2023},
  publisher={IEEE}
}

@article{chen2021infrared,
  title={Infrared action detection in the dark via cross-stream attention mechanism},
  author={Chen, Xu and Gao, Chenqiang and Li, Chaoyu and Yang, Yi and Meng, Deyu},
  journal={IEEE Transactions on Multimedia},
  volume={24},
  pages={288--300},
  year={2022},
  doi={10.1109/TMM.2021.3050069},
  publisher={IEEE}
}

@misc{flir2021,
  author = {F. Team and others},
  title = {Free FLIR Thermal Dataset for Algorithm Training},
  year = {2021},
  url = {https://www.flir.com/oem/adas/adas-dataset-form/}
}

@inproceedings{jia2021llvip,
  title={LLVIP: A visible-infrared paired dataset for low-light vision},
  author={Jia, Xinyu and Zhu, Chuang and Li, Minzhen and Tang, Wenqi and Zhou, Wenli},
  booktitle={Proceedings of the IEEE/CVF international conference on computer vision},
  pages={3496--3504},
  year={2021}
}

@article{liu2015general,
  title={A general framework for image fusion based on multi-scale transform and sparse representation},
  author={Liu, Yu and Liu, Shuping and Wang, Zengfu},
  journal={Information fusion},
  volume={24},
  pages={147--164},
  year={2015},
  publisher={Elsevier}
}

@article{ma2016infrared,
  title={Infrared and visible image fusion via gradient transfer and total variation minimization},
  author={Ma, Jiayi and Chen, Chen and Li, Chang and Huang, Jun},
  journal={Information Fusion},
  volume={31},
  pages={100--109},
  year={2016},
  publisher={Elsevier}
}

@article{li2020mdlatlrr,
  title={MDLatLRR: A novel decomposition method for infrared and visible image fusion},
  author={Li, Hui and Wu, Xiao-Jun and Kittler, Josef},
  journal={IEEE Transactions on Image Processing},
  volume={29},
  pages={4733--4746},
  year={2020},
  publisher={IEEE}
}

@article{li2018densefuse,
  title={DenseFuse: A fusion approach to infrared and visible images},
  author={Li, Hui and Wu, Xiao-Jun},
  journal={IEEE Transactions on Image Processing},
  volume={28},
  number={5},
  pages={2614--2623},
  year={2019},
  doi={10.1109/TIP.2018.2887342},
  publisher={IEEE}
}

@article{tang2022piafusion,
  title={PIAFusion: A progressive infrared and visible image fusion network based on illumination aware},
  author={Tang, Linfeng and Yuan, Jiteng and Zhang, Hao and Jiang, Xingyu and Ma, Jiayi},
  journal={Information Fusion},
  volume={83},
  pages={79--92},
  year={2022},
  publisher={Elsevier}
}

@article{ma2019fusiongan,
  title={FusionGAN: A generative adversarial network for infrared and visible image fusion},
  author={Ma, Jiayi and Yu, Wei and Liang, Pengwei and Li, Chang and Jiang, Junjun},
  journal={Information fusion},
  volume={48},
  pages={11--26},
  year={2019},
  publisher={Elsevier}
}

@inproceedings{liu2022target,
  title={Target-aware dual adversarial learning and a multi-scenario multi-modality benchmark to fuse infrared and visible for object detection},
  author={Liu, Jinyuan and Fan, Xin and Huang, Zhanbo and Wu, Guanyao and Liu, Risheng and Zhong, Wei and Luo, Zhongxuan},
  booktitle={Proceedings of the IEEE/CVF conference on computer vision and pattern recognition},
  pages={5802--5811},
  year={2022}
}

@article{ma2022swinfusion,
  title={SwinFusion: Cross-domain long-range learning for general image fusion via swin transformer},
  author={Ma, Jiayi and Tang, Linfeng and Fan, Fan and Huang, Jun and Mei, Xiaoguang and Ma, Yong},
  journal={IEEE/CAA Journal of Automatica Sinica},
  volume={9},
  number={7},
  pages={1200--1217},
  year={2022},
  publisher={IEEE}
}

@article{xu2020u2fusion,
  title={U2Fusion: A unified unsupervised image fusion network},
  author={Xu, Han and Ma, Jiayi and Jiang, Junjun and Guo, Xiaojie and Ling, Haibin},
  journal={IEEE Transactions on Pattern Analysis and Machine Intelligence},
  volume={44},
  number={1},
  pages={502--518},
  year={2022},
  doi={10.1109/TPAMI.2020.3012548},
  publisher={IEEE}
}

@inproceedings{zhang2021guided,
  title={Guided attentive feature fusion for multispectral pedestrian detection},
  author={Zhang, Heng and Fromont, Elisa and Lef{\`e}vre, S{\'e}bastien and Avignon, Bruno},
  booktitle={Proceedings of the IEEE/CVF winter conference on applications of computer vision},
  pages={72--80},
  year={2021}
}

@inproceedings{dhanaraj2020vehicle,
  title={Vehicle detection from multi-modal aerial imagery using YOLOv3 with mid-level fusion},
  author={Dhanaraj, Mayur and Sharma, Manish and Sarkar, Tiyasa and Karnam, Srivallabha and Chachlakis, Dimitris G and Ptucha, Raymond and Markopoulos, Panos P and Saber, Eli},
  booktitle={Big data II: learning, analytics, and applications},
  volume={11395},
  pages={22--32},
  year={2020},
  organization={SPIE}
}

@inproceedings{zhang2020multispectral,
  title={Multispectral fusion for object detection with cyclic fuse-and-refine blocks},
  author={Zhang, Heng and Fromont, Elisa and Lefevre, S{\'e}bastien and Avignon, Bruno},
  booktitle={2020 IEEE International conference on image processing (ICIP)},
  pages={276--280},
  year={2020},
  organization={IEEE}
}

@article{qingyun2022cross,
  title={Cross-modality attentive feature fusion for object detection in multispectral remote sensing imagery},
  author={Qingyun, Fang and Zhaokui, Wang},
  journal={Pattern Recognition},
  volume={130},
  pages={108786},
  year={2022},
  publisher={Elsevier}
}

@article{fu2023lraf,
  title={Lraf-net: Long-range attention fusion network for visible--infrared object detection},
  author={Fu, Haolong and Wang, Shixun and Duan, Puhong and Xiao, Changyan and Dian, Renwei and Li, Shutao and Li, Zhiyong},
  journal={IEEE Transactions on Neural Networks and Learning Systems},
  volume={35},
  number={10},
  pages={13232--13245},
  year={2024},
  doi={10.1109/TNNLS.2023.3266452},
  publisher={IEEE}
}

@inproceedings{liu2023multi,
  title={Multi-interactive feature learning and a full-time multi-modality benchmark for image fusion and segmentation},
  author={Liu, Jinyuan and Liu, Zhu and Wu, Guanyao and Ma, Long and Liu, Risheng and Zhong, Wei and Luo, Zhongxuan and Fan, Xin},
  booktitle={Proceedings of the IEEE/CVF international conference on computer vision},
  pages={8115--8124},
  year={2023}
}

@misc{yolov5,
  title        = {{YOLOv5} by Ultralytics},
  author       = {Jocher, Glenn and Chaurasia, Ayush and Borovec, Jirka and others},
  year         = {2020},
  howpublished = {\url{https://github.com/ultralytics/yolov5}}
}

@inproceedings{zhao2023cddfuse,
  title={Cddfuse: Correlation-driven dual-branch feature decomposition for multi-modality image fusion},
  author={Zhao, Zixiang and Bai, Haowen and Zhang, Jiangshe and Zhang, Yulun and Xu, Shuang and Lin, Zudi and Timofte, Radu and Van Gool, Luc},
  booktitle={Proceedings of the IEEE/CVF conference on computer vision and pattern recognition},
  pages={5906--5916},
  year={2023}
}

@article{wang2025multi,
  title={Multi-text guidance is important: Multi-modality image fusion via large generative vision-language model},
  author={Wang, Zeyu and Zhao, Libo and Zhang, Jizheng and Song, Rui and Song, Haiyu and Meng, Jiana and Wang, Shidong},
  journal={International Journal of Computer Vision},
  volume={133},
  number={7},
  pages={4646--4668},
  year={2025},
  publisher={Springer}
}

@inproceedings{yi2024text,
  title={Text-if: Leveraging semantic text guidance for degradation-aware and interactive image fusion},
  author={Yi, Xunpeng and Xu, Han and Zhang, Hao and Tang, Linfeng and Ma, Jiayi},
  booktitle={Proceedings of the IEEE/CVF Conference on Computer Vision and Pattern Recognition},
  pages={27026--27035},
  year={2024}
}

@inproceedings{zhang2023dense,
  title={Dense distinct query for end-to-end object detection},
  author={Zhang, Shilong and Wang, Xinjiang and Wang, Jiaqi and Pang, Jiangmiao and Lyu, Chengqi and Zhang, Wenwei and Luo, Ping and Chen, Kai},
  booktitle={Proceedings of the IEEE/CVF conference on computer vision and pattern recognition},
  pages={7329--7338},
  year={2023}
}

@inproceedings{zhao2023metafusion,
  title={Metafusion: Infrared and visible image fusion via meta-feature embedding from object detection},
  author={Zhao, Wenda and Xie, Shigeng and Zhao, Fan and He, You and Lu, Huchuan},
  booktitle={Proceedings of the IEEE/CVF Conference on Computer Vision and Pattern Recognition},
  pages={13955--13965},
  year={2023}
}

@inproceedings{chen2022multimodal,
  title={Multimodal object detection via probabilistic ensembling},
  author={Chen, Yi-Ting and Shi, Jinghao and Ye, Zelin and Mertz, Christoph and Ramanan, Deva and Kong, Shu},
  booktitle={European Conference on Computer Vision},
  pages={139--158},
  year={2022},
  organization={Springer}
}

@article{tang2023datfuse,
  title={DATFuse: Infrared and visible image fusion via dual attention transformer},
  author={Tang, Wei and He, Fazhi and Liu, Yu and Duan, Yansong and Si, Tongzhen},
  journal={IEEE Transactions on Circuits and Systems for Video Technology},
  volume={33},
  number={7},
  pages={3159--3172},
  year={2023},
  publisher={IEEE}
}

@inproceedings{cao2023multimodal,
  title={Multimodal object detection by channel switching and spatial attention},
  author={Cao, Yue and Bin, Junchi and Hamari, Jozsef and Blasch, Erik and Liu, Zheng},
  booktitle={Proceedings of the IEEE/CVF Conference on Computer Vision and Pattern Recognition},
  pages={403--411},
  year={2023}
}

@article{zeng2024mmi,
  title={MMI-Det: Exploring multi-modal integration for visible and infrared object detection},
  author={Zeng, Yuqiao and Liang, Tengfei and Jin, Yi and Li, Yidong},
  journal={IEEE Transactions on Circuits and Systems for Video Technology},
  volume={34},
  number={11},
  pages={11198--11213},
  year={2024},
  doi={10.1109/TCSVT.2024.3418965},
  publisher={IEEE}
}

@inproceedings{he2023multispectral,
  title={Multispectral object detection via cross-modal conflict-aware learning},
  author={He, Xiao and Tang, Chang and Zou, Xin and Zhang, Wei},
  booktitle={Proceedings of the 31st ACM International Conference on Multimedia},
  pages={1465--1474},
  year={2023}
}

@article{lu2024gan,
  title={GAN-HA: A generative adversarial network with a novel heterogeneous dual-discriminator network and a new attention-based fusion strategy for infrared and visible image fusion},
  author={Lu, Guosheng and Fang, Zile and Tian, Jiaju and Huang, Haowen and Xu, Yuelong and Han, Zhuolin and Kang, Yaoming and Feng, Can and Zhao, Zhigang},
  journal={Infrared Physics \& Technology},
  volume={142},
  pages={105548},
  year={2024},
  publisher={Elsevier}
}

@inproceedings{li2025fd2,
  title={FD2-Net: Frequency-driven feature decomposition network for infrared-visible object detection},
  author={Li, Ke and Wang, Di and Hu, Zhangyuan and Li, Shaofeng and Ni, Weiping and Zhao, Lin and Wang, Quan},
  booktitle={Proceedings of the AAAI Conference on Artificial Intelligence},
  volume={39},
  number={5},
  pages={4797--4805},
  year={2025}
}

@article{dong2024fusion,
  title={Fusion-mamba for cross-modality object detection},
  author={Dong, Wenhao and Zhu, Haodong and Lin, Shaohui and Luo, Xiaoyan and Shen, Yunhang and Liu, Xuhui and Zhang, Juan and Guo, Guodong and Zhang, Baochang},
  journal={arXiv preprint arXiv:2404.09146},
  year={2024}
}

@article{shen2024icafusion,
  title={ICAFusion: Iterative cross-attention guided feature fusion for multispectral object detection},
  author={Shen, Jifeng and Chen, Yifei and Liu, Yue and Zuo, Xin and Fan, Heng and Yang, Wankou},
  journal={Pattern Recognition},
  volume={145},
  pages={109913},
  year={2024},
  publisher={Elsevier}
}

@article{yang2024multidimensional,
  title={Multidimensional Fusion Network for Multispectral Object Detection},
  author={Yang, Fan and Liang, Binbin and Li, Wei and Zhang, Jianwei},
  journal={IEEE Transactions on Circuits and Systems for Video Technology},
  volume={35},
  number={1},
  pages={547--560},
  year={2025},
  doi={10.1109/TCSVT.2024.3454631},
  publisher={IEEE}
}

@article{tang2024itfuse,
  title={ITFuse: An interactive transformer for infrared and visible image fusion},
  author={Tang, Wei and He, Fazhi and Liu, Yu},
  journal={Pattern Recognition},
  volume={156},
  pages={110822},
  year={2024},
  publisher={Elsevier}
}

@article{tang2022image,
  title={Image fusion in the loop of high-level vision tasks: A semantic-aware real-time infrared and visible image fusion network},
  author={Tang, Linfeng and Yuan, Jiteng and Ma, Jiayi},
  journal={Information Fusion},
  volume={82},
  pages={28--42},
  year={2022},
  publisher={Elsevier}
}

@article{wang2024efficient,
  title={An efficient frequency domain fusion network of infrared and visible images},
  author={Wang, Chenwu and Wu, Junsheng and Fang, Aiqing and Zhu, Zhixiang and Wang, Pei and Chen, Hao},
  journal={Engineering Applications of Artificial Intelligence},
  volume={133},
  pages={108013},
  year={2024},
  publisher={Elsevier}
}

@article{yue2023dif,
  title={Dif-fusion: Toward high color fidelity in infrared and visible image fusion with diffusion models},
  author={Yue, Jun and Fang, Leyuan and Xia, Shaobo and Deng, Yue and Ma, Jiayi},
  journal={IEEE Transactions on Image Processing},
  volume={32},
  pages={5705--5720},
  year={2023},
  publisher={IEEE}
}

@article{wang2025rffusion,
  title={Efficient Rectified Flow for Image Fusion},
  author={Wang, Zirui and Zhang, Jiayi and Guan, Tianwei and Zhou, Yuhan and Li, Xingyuan and Dong, Minjing and Liu, Jinyuan},
  journal={arXiv preprint arXiv:2509.16549},
  year={2025}
}

@inproceedings{dang2025ddfd,
  title={DDFD: Diffusion-Based Denoising Fusion for Object Detection in Infrared-Visible Images},
  author={Dang, Min and Liu, Gang and Zhao, Jingqi and Kong, Adams Wai-Kin and Luo, Nan and Wang, Di},
  booktitle={Proceedings of the 33rd ACM International Conference on Multimedia},
  pages={1452--1461},
  year={2025}
}

@inproceedings{yang_customizedfusion,
  title={Customized Fusion: A Closed-Loop Dynamic Network for Adaptive Multi-Task-Aware Infrared-Visible Image Fusion},
  author={Yang, Zengyi and Liu, Yu and Cheng, Juan and Zhu, Zhiqin and Zhang, Yafei and Li, Huafeng},
  booktitle={Proceedings of the IEEE/CVF Conference on Computer Vision and Pattern Recognition},
  pages={188--198},
  year={2026},
  url={https://github.com/YR0211/CLDyN}
}

@article{bao2023heat,
  title={Heat-assisted detection and ranging},
  author={Bao, Fanglin and Wang, Xueji and Sureshbabu, Shree Hari and Sreekumar, Gautam and Yang, Liping and Aggarwal, Vaneet and Boddeti, Vishnu N. and Jacob, Zubin},
  journal={Nature},
  volume={619},
  number={7971},
  pages={743--748},
  year={2023}
}

@inproceedings{yuan_vtintrinsic,
  title={{VT-Intrinsic}: Physics-Based Decomposition of Reflectance and Shading using a Single Visible-Thermal Image Pair},
  author={Yuan, Zeqing and Ramanagopal, Mani and Sankaranarayanan, Aswin C. and Narasimhan, Srinivasa G.},
  booktitle={Proceedings of the IEEE/CVF Conference on Computer Vision and Pattern Recognition},
  pages={41752--41761},
  year={2026},
  url={https://vt-intrinsic.github.io}
}

@inproceedings{gao_physirsplat,
  title={{PhysIR-Splat}: Physically Consistent Thermal Infrared Radiative Transfer in {3D} Gaussian Splatting},
  author={Gao, Jingyuan and Hu, Yumeng and Gao, Fei and Zhang, Mingjin},
  booktitle={Proceedings of the IEEE/CVF Conference on Computer Vision and Pattern Recognition},
  pages={11818--11828},
  year={2026},
  url={https://github.com/JingyuanGao0919/physir-splat}
}

@inproceedings{jiang_phyfusion,
  title={{PhyFusion}: Physics-Aware Infrared and Visible Image Fusion via Modality-Specific Physical Priors},
  author={Jiang, Haiyang and Zhang, Huiqin and Zhang, Yanduo and Ma, Jiayi and Jiang, Junjun and Zhou, Huabing},
  booktitle={Proceedings of the IEEE/CVF Conference on Computer Vision and Pattern Recognition Findings},
  pages={4993--5002},
  year={2026}
}

@inproceedings{li2026mvrmoe,
author = {Li, Daoheng and Qian, Jianjun and Yan, Mengkai and Yang, Jian},
title = {Multi-View Structural Representations for Physically-Consistent TeX Generation},
year = {2026},
isbn = {9798400722134},
publisher = {Association for Computing Machinery},
address = {New York, NY, USA},
url = {https://doi.org/10.1145/3767308.3835251},
doi = {10.1145/3767308.3835251},
booktitle = {Proceedings of the 34th ACM International Conference on Multimedia},
pages = {4416–4425},
numpages = {10},
location = {Rio de Janeiro, Brazil},
series = {MM '26}
}

@article{li2025tdhf,
  title={{TDHF}: Task-Driven Hierarchical Image Fusion Under Low-Light Conditions},
  author={Li, Daoheng and Yan, Mengkai and Li, Jun and Yang, Jian and Qian, Jianjun},
  journal={IEEE Transactions on Circuits and Systems for Video Technology},
  year={2026},
  volume={36},
  number={4},
  pages={4064--4077},
  doi={10.1109/TCSVT.2025.3635231},
  publisher={IEEE}
}

@article{zhu2024hyper,
  title={Hyper-connections},
  author={Zhu, Defa and Huang, Hongzhi and Huang, Zihao and Zeng, Yutao and Mao, Yunyao and Wu, Banggu and Min, Qiyang and Zhou, Xun},
  journal={arXiv preprint arXiv:2409.19606},
  year={2024}
}

@inproceedings{zamir2022restormer,
  title={{Restormer}: Efficient Transformer for High-Resolution Image Restoration},
  author={Zamir, Syed Waqas and Arora, Aditya and Khan, Salman and Hayat, Munawar and Khan, Fahad Shahbaz and Yang, Ming-Hsuan},
  booktitle={Proceedings of the IEEE/CVF Conference on Computer Vision and Pattern Recognition},
  year={2022},
  pages={5728--5739}
}

@article{hwang2025multispectral,
  title={Multispectral Detection Transformer with Infrared-Centric Feature Fusion},
  author={Hwang, Seongmin and Han, Daeyoung and Jeon, Moongu},
  journal={arXiv preprint arXiv:2505.15137},
  year={2025},
  eprint={2505.15137},
  archivePrefix={arXiv},
  primaryClass={cs.CV},
  url={https://arxiv.org/abs/2505.15137},
  note={Version 2; under review}
}

@inproceedings{zhu2025wavemamba,
  title={{WaveMamba}: Wavelet-Driven {Mamba} Fusion for {RGB}-Infrared Object Detection},
  author={Zhu, Haodong and Dong, Wenhao and Yang, Linlin and Li, Hong and Yang, Yuguang and Ren, Yangyang and Zhu, Qingcheng and Feng, Zichao and Li, Changbai and Lin, Shaohui and Wang, Runqi and Luo, Xiaoyan and Zhang, Baochang},
  booktitle={Proceedings of the IEEE/CVF International Conference on Computer Vision},
  pages={11219--11229},
  year={2025},
  doi={10.1109/ICCV51701.2025.01044},
  eprint={2507.18173},
  archivePrefix={arXiv},
  primaryClass={cs.CV},
  url={https://openaccess.thecvf.com/content/ICCV2025/html/Zhu_WaveMamba_Wavelet-Driven_Mamba_Fusion_for_RGB-Infrared_Object_Detection_ICCV_2025_paper.html}
}

@misc{wang2026registerbridgemm,
  title={{RegisterBridgeMM}: A Register-Centric Framework for {RGB}-Infrared Object Detection},
  author={Wang, Zian and Liang, Hangchuan and Chen, Yuehua and Li, Changchun and Guo, Chaoyi and Liu, Mingzhe and Gu, Fangming},
  year={2026},
  eprint={2608.04833},
  archivePrefix={arXiv},
  primaryClass={cs.CV},
  url={https://arxiv.org/abs/2608.04833}
}

@misc{feng2026infranet,
  title={{InfraNet}: Quality-Aware {RGB} Guidance for Efficient Infrared Object Detection},
  author={Feng, Zichao and Zhu, Haodong and Yang, Jingying and Xu, Sheng and Ren, Yangyang and Yang, Yuguang and Liu, Xuhui and Zhang, Juan and Wang, Tian and Yang, Linlin and Zhang, Baochang},
  year={2026},
  eprint={2607.03795},
  archivePrefix={arXiv},
  primaryClass={cs.CV},
  url={https://arxiv.org/abs/2607.03795}
}

@inproceedings{hao2026distribution,
  title={Distribution-Aligned Multimodal Fusion for Robust Object Detection},
  author={Hao, Xiaohui and Pu, Yanglin and Wang, Yongjun and She, Rui},
  booktitle={Proceedings of the IEEE/CVF Conference on Computer Vision and Pattern Recognition},
  pages={25494--25503},
  month={June},
  year={2026}
}

@article{xiang2026hmal,
  title={Learning Modality Advantage Hierarchically for {RGB}-Infrared Object Detection},
  author={Xiang, Zelin and Li, Xuwei},
  journal={IET Image Processing},
  volume={20},
  number={1},
  year={2026},
  note={Article e70394},
  doi={10.1049/ipr2.70394}
}

@article{wu2026fredft,
  title={{FreDFT}: Frequency Domain Fusion Transformer for Visible-Infrared Object Detection},
  author={Wu, Wencong and Zhang, Xiuwei and Yin, Hanlin and Dai, Shun and Zhang, Hongxi and Zhang, Yanning},
  journal={IEEE Transactions on Multimedia},
  year={2026},
  note={Early access},
  doi={10.1109/TMM.2026.3712630}
}

@article{lu2025illumination,
  title={Illumination-Aware Multimodal Hierarchical Fusion Network for {RGB}-Infrared Object Detection},
  author={Lu, Ting and Lu, Jiacheng and Fu, Wei and Xi, Yifan},
  journal={IEEE Transactions on Geoscience and Remote Sensing},
  volume={63},
  pages={1--14},
  year={2025},
  doi={10.1109/TGRS.2025.3636590}
}

@article{sun2022drone,
  title={Drone-Based {RGB}-Infrared Cross-Modality Vehicle Detection Via Uncertainty-Aware Learning},
  author={Sun, Yiming and Cao, Bing and Zhu, Pengfei and Hu, Qinghua},
  journal={IEEE Transactions on Circuits and Systems for Video Technology},
  volume={32},
  number={10},
  pages={6700--6713},
  year={2022},
  doi={10.1109/TCSVT.2022.3168279}
}

@article{hou2026chudetr,
  title={Infrared-Visible Dual-Modal Object Detection Transformer with Cross-Hierarchical Attention and {U}-Like Gated Interaction},
  author={Hou, Zhiqiang and Wu, Xingping and Zhao, Jialu and Ma, Sugang and Liu, Yang and Lu, Ruitao and Xi, Jianxiang},
  journal={Infrared Physics \& Technology},
  volume={158},
  pages={106787},
  year={2026},
  doi={10.1016/j.infrared.2026.106787}
}

@inproceedings{tdfusion2025,
  title={Task-driven Image Fusion with Learnable Fusion Loss},
  author={Bai, Haowen and Zhang, Jiangshe and Zhao, Zixiang and Wu, Yichen and Deng, Lilun and Cui, Yukun and Feng, Tao and Xu, Shuang},
  booktitle={Proceedings of the IEEE/CVF Conference on Computer Vision and Pattern Recognition},
  pages={7457--7468},
  year={2025},
  doi={10.1109/CVPR52734.2025.00699}
}

@inproceedings{zhao2024emma,
  title={Equivariant Multi-Modality Image Fusion},
  author={Zhao, Zixiang and Bai, Haowen and Zhang, Jiangshe and Zhang, Yulun and Zhang, Kai and Xu, Shuang and Chen, Dongdong and Timofte, Radu and Van Gool, Luc},
  booktitle={Proceedings of the IEEE/CVF Conference on Computer Vision and Pattern Recognition},
  pages={25912--25921},
  year={2024},
  doi={10.1109/CVPR52733.2024.02448}
}

@article{wang2024dcinn,
  title={A General Paradigm with Detail-Preserving Conditional Invertible Network for Image Fusion},
  author={Wang, Wu and Deng, Liang-Jian and Ran, Ran and Vivone, Gemine},
  journal={International Journal of Computer Vision},
  volume={132},
  number={4},
  pages={1029--1054},
  year={2024},
  doi={10.1007/s11263-023-01924-5}
}

@inproceedings{zhang2024mrfs,
  title={{MRFS}: Mutually Reinforcing Image Fusion and Segmentation},
  author={Zhang, Hao and Zuo, Xuhui and Jiang, Jie and Guo, Chunchao and Ma, Jiayi},
  booktitle={Proceedings of the IEEE/CVF Conference on Computer Vision and Pattern Recognition},
  pages={26964--26973},
  year={2024},
  doi={10.1109/CVPR52733.2024.02547}
}

@article{liu2024timfusion,
  title={A Task-Guided, Implicitly-Searched and Meta-Initialized Deep Model for Image Fusion},
  author={Liu, Risheng and Liu, Zhu and Liu, Jinyuan and Fan, Xin and Luo, Zhongxuan},
  journal={IEEE Transactions on Pattern Analysis and Machine Intelligence},
  volume={46},
  number={10},
  pages={6594--6609},
  year={2024},
  doi={10.1109/TPAMI.2024.3382308}
}

@article{yang2025kdfuse,
  title={{KDFuse}: A High-Level Vision Task-Driven Infrared and Visible Image Fusion Method Based on Cross-Domain Knowledge Distillation},
  author={Yang, Chenjia and Luo, Xiaoqing and Zhang, Zhancheng and Chen, Zhiguo and Wu, Xiao-Jun},
  journal={Information Fusion},
  volume={118},
  pages={102944},
  year={2025},
  doi={10.1016/j.inffus.2025.102944}
}

@inproceedings{cheng2025gifnet,
  title={One Model for {ALL}: Low-Level Task Interaction Is a Key to Task-Agnostic Image Fusion},
  author={Cheng, Chunyang and Xu, Tianyang and Feng, Zhenhua and Wu, Xiaojun and Tang, Zhangyong and Li, Hui and Zhang, Zeyang and Atito, Sara and Awais, Muhammad and Kittler, Josef},
  booktitle={Proceedings of the IEEE/CVF Conference on Computer Vision and Pattern Recognition},
  pages={28102--28112},
  year={2025},
  doi={10.1109/CVPR52734.2025.02617}
}

@article{wang2025dafusion,
  title={A Degradation-Aware Guided Fusion Network for Infrared and Visible Image},
  author={Wang, Xue and Guan, Zheng and Qian, Wenhua and Cao, Jinde and Ma, Runzhuo and Bi, Cong},
  journal={Information Fusion},
  volume={118},
  pages={102931},
  year={2025},
  doi={10.1016/j.inffus.2025.102931}
}

@inproceedings{wu2025sage,
  title={Every {SAM} Drop Counts: Embracing Semantic Priors for Multi-Modality Image Fusion and Beyond},
  author={Wu, Guanyao and Liu, Haoyu and Fu, Hongming and Peng, Yichuan and Liu, Jinyuan and Fan, Xin and Liu, Risheng},
  booktitle={Proceedings of the IEEE/CVF Conference on Computer Vision and Pattern Recognition},
  pages={17882--17891},
  year={2025},
  doi={10.1109/CVPR52734.2025.01666}
}

@article{cheng2025fusionbooster,
  title={{FusionBooster}: A Unified Image Fusion Boosting Paradigm},
  author={Cheng, Chunyang and Xu, Tianyang and Wu, Xiao-Jun and Li, Hui and Li, Xi and Kittler, Josef},
  journal={International Journal of Computer Vision},
  volume={133},
  number={5},
  pages={3041--3058},
  year={2025},
  doi={10.1007/s11263-024-02266-6}
}

@article{he2026missfusion,
  title={{MISSFusion}: A Multi-Granularity Interaction and Semantic-Structure Rectification Network for Infrared and Visible Image Fusion},
  author={He, Yuanyuan and Gao, Xueyan and Liu, Rui and Zhou, Yadong and Zhou, Chengjiang and Zhao, Shan and Yang, Yang},
  journal={Journal of King Saud University Computer and Information Sciences},
  volume={38},
  number={7},
  pages={790},
  year={2026},
  doi={10.1007/s44443-026-01122-6}
}

@article{zhao2025freefusion,
  author={Zhao, Wenda and Cui, Hengshuai and Wang, Haipeng and He, You and Lu, Huchuan},
  title={{FreeFusion}: Infrared and Visible Image Fusion via Cross Reconstruction Learning},
  journal={IEEE Transactions on Pattern Analysis and Machine Intelligence},
  volume={47},
  number={9},
  pages={8040--8056},
  year={2025},
  doi={10.1109/TPAMI.2025.3572599}
}

@article{zhao2025cdtfusion,
  title={CDTFusion: Crossing domain and task for infrared and visible image fusion},
  author={Zhao, Wenda and Wang, Wenbo and Wang, Haipeng and He, You and Lu, Huchuan},
  journal={IEEE Transactions on Pattern Analysis and Machine Intelligence},
  year={2025},
  publisher={IEEE}
}

\begin{IEEEbiography}[{\includegraphics[width=1in,height=1.25in,clip,keepaspectratio]{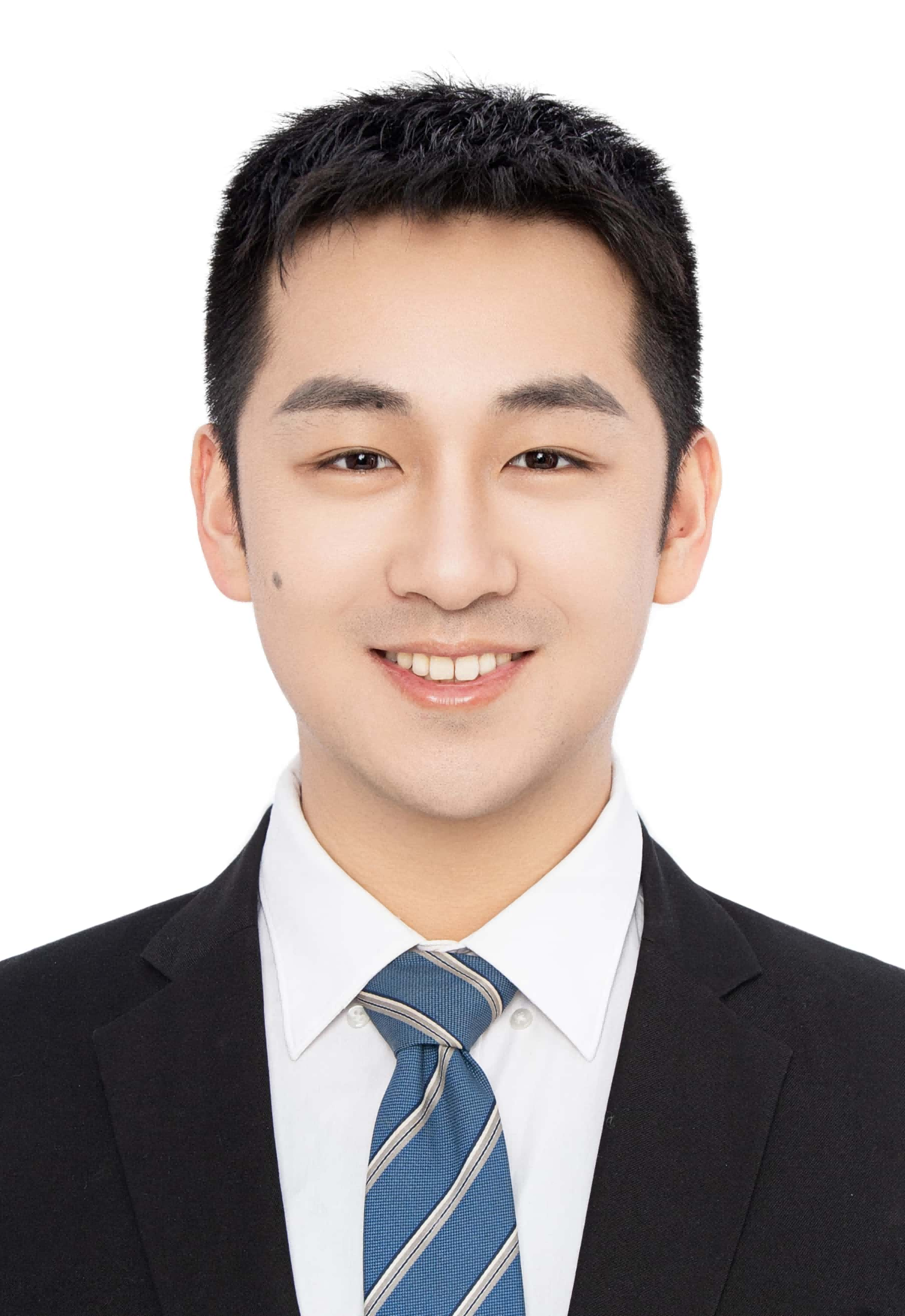}}]
{Daoheng Li} received the M.S. degree from Shandong Jianzhu University, Jinan, China, in 2023. 
He is currently pursuing the Ph.D. degree with the School of Computer Science and Engineering, Nanjing University of Science and Technology, Nanjing, China. 
His research interests include computer vision and pattern recognition, with a particular focus on low-light visual perception and thermal infrared image analysis.
\end{IEEEbiography}



\begin{IEEEbiography}[{\includegraphics[width=1in,height=1.25in,clip,keepaspectratio]{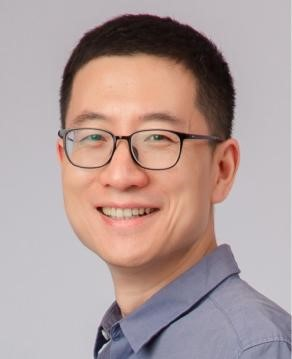}}]
{Jianjun Qian} received the PhD degree from Nanjing University of Science and Technology (NUST), on the subject of pattern recognition and intelligence systems in 2014. He is currently a professor at the School of Computer Science and Engineering of NUST. He was selected as a Hong Kong Scholar in China in 2018. From 2019 to 2021, he was a Postdoctoral Fellow at ITC of Hong Kong Polytechnic University. He has published about 100 top journal and conference papers, such as TPAMI/TIP/TNNLS, IJCV, CVPR, ICCV, AAAI, ECCV, etc. His current research interests include pattern recognition, computer vision, especially in human-centric visual computing.
\end{IEEEbiography}

\begin{IEEEbiography}[{\includegraphics[width=1in,height=1.25in,clip,keepaspectratio]{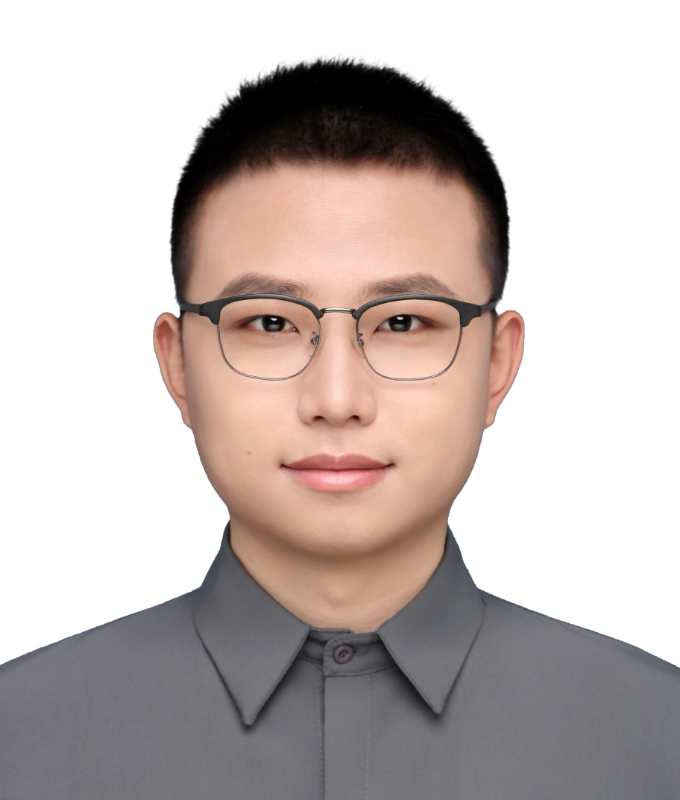}}]
{Mengkai Yan} received his Ph.D. in Control Science and Engineering from Nanjing University of Science and Technology in 2025, with a research focus on computer vision. 
He is currently a lecturer at Hohai University. His current research interests include computer vision and anomaly detection.
\end{IEEEbiography}

\begin{IEEEbiography}[{\includegraphics[width=1in,height=1.25in,clip,keepaspectratio]{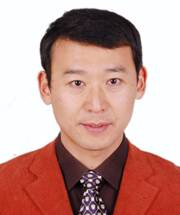}}]
{Jian Yang} received the PhD degree from Nanjing University of Science and Technology (NUST) in 2002, majoring in pattern recognition and intelligence systems. From 2003 to 2007, he was a Postdoctoral Fellow at the University of Zaragoza, Hong Kong Polytechnic University and New Jersey Institute of Technology, respectively. From 2007 to present, he is a professor in the School of Computer Science and Technology of NUST, and a visiting professor of Nanjing University. His papers have been cited over 60000 times in the Scholar Google. His research interests include pattern recognition and computer vision. Currently, he is/was an associate editor of Pattern Recognition, Pattern Recognition Letters, IEEE Trans. Neural Networks and Learning Systems, and Neurocomputing. He is a Fellow of IAPR.
\end{IEEEbiography}

\end{document}